\documentclass{article}
\usepackage{enumitem}
\usepackage{iclr2027_conference,times}

\usepackage{amsmath,amsfonts,bm}

\def\eqref#1{equation~\ref{#1}}

\def\1{\bm{1}}

\DeclareMathAlphabet{\mathsfit}{\encodingdefault}{\sfdefault}{m}{sl}
\SetMathAlphabet{\mathsfit}{bold}{\encodingdefault}{\sfdefault}{bx}{n}

\usepackage{hyperref}
\usepackage{url}
\usepackage{graphicx} 
\usepackage{subcaption}
\usepackage{natbib}
\usepackage{booktabs} 
\usepackage{float}
\usepackage{algorithm}
\usepackage{algpseudocode}
\newcommand{\method}{ColorTuning }
\usepackage[normalem]{ulem}
\newsavebox{\gcbfullbox}
\usepackage{xcolor}

\title{On Color Alignment in VAE Latent Spaces \\ and Its Applications}

\author{
\textbf{Julian D. Santamaria}$^{1,2,}$, \textbf{Kai Wang}$^{3}$, \textbf{Jesús Malo}$^{4}$, \\  \vspace{0.05cm} 
\textbf{ Javier Vazquez-Corral}$^{1,2}$, \textbf{Alexandra Gómez-Villa}$^{1,2}$ \\
$^{1}$Computer Vision Center, $^{2}$Universitat Autònoma de Barcelona, \\ $^{3}$City University of Hong Kong (Dongguan), \vspace{0.05cm} $^{4}$Universitat de València \\ 
\texttt{jsantamaria@cvc.uab.cat}, \texttt{kai.wang@cityu-dg.edu.cn}, \\ 
\texttt{jesus.malo@uv.es}, \texttt{\{jvazquez, agomezvi\}@cvc.uab.cat}
}

\iclrfinalcopy 
\begin{document}

\maketitle

\begin{figure}[h]
\begin{center}

\includegraphics[width=0.95\linewidth]{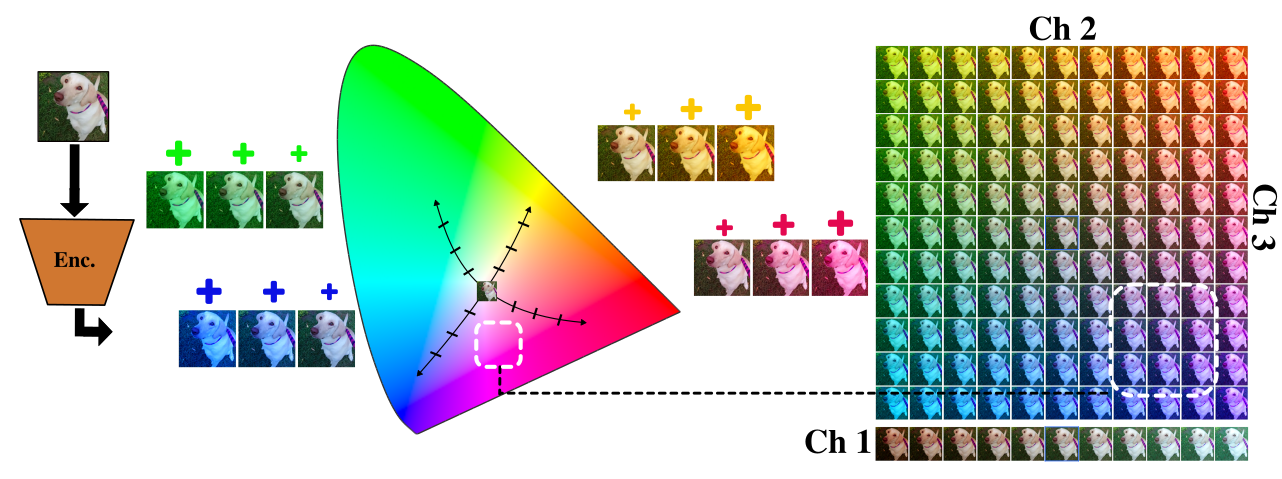}
\end{center}
\caption{A color space hides in plain sight in the latent space of an autoencoder. Train a VAE with three latent channels on natural images, encode an image (left), add a constant offset $\delta$ to one channel, and decode: the image changes color while preserving its content. Sweeping two channels (right; ch2 across columns, ch3 across rows) spans an opponent-color plane and all intermediate mixtures, while the remaining channel (bottom row) controls brightness.}

\label{fig:teaser}
\end{figure}

\begin{abstract}
Variational autoencoders (VAEs) are a key part of modern text-to-image models, which generate images within their latent space. VAEs are known to disentangle the main factors of variation in the data, and color is known to be one of the most structured of these in natural images: decorrelating it yields one luminance axis and two opponent-color axes. Color should therefore be expected to emerge as a distinct factor in the VAE latent space. Yet how these latent spaces represent color remains largely unexplored. In this work, we show that the VAEs of text-to-image models share a color subspace aligned with brightness and opponent-colors. Through a linear approximation of the encoder and targeted latent steering, we find this subspace consistently across a broad range of VAEs, from SD1.5 to FLUX.2 and Z-Image. Building on this characterization, we propose three applications: \emph{\method}, which achieves state-of-the-art 
in precise numerical color generation on the fine-grained CSS3/X11 system of GenColorBench, \emph{saturation control}, to adjust the global chromatic intensity, and \emph{color transfer}, to change the palette to match a reference. The code and models are publicly available at \url{https://julian075.github.io/Color_Subspace/}
\end{abstract}
\newpage

\section{Introduction}\label{sec_introduction}

Natural images are far from random: they occupy a small, highly structured subset of the space of all possible pixel arrays, with strong spatial and spectral correlations 
\citep{simoncelli2001natural}. Efficient coding theory holds that a visual system should exploit this redundancy by mapping its inputs to a compact, decorrelated representation \citep{barlow1961possible,Malo20}. For color, this principle gives a remarkably
consistent answer: decorrelating the cone responses to natural scenes yields one achromatic luminance axis and two chromatic opponent axes, namely red--green and blue--yellow \citep{buchsbaum1983trichromacy,ruderman1998statistics}. These axes closely match the cardinal directions of early human color vision \citep{derrington1984chromatic} and are mirrored in perceptual color spaces such as CIELAB. Luminance and opponent colors are therefore not an arbitrary convention but a property of natural images, one that any efficient
encoder should be expected to recover.

In fact, the evidence suggests that artificial systems trained on natural images develop this color representation too. Unsupervised methods related to independent component analysis, applied to color image patches, learn separate achromatic and color-opponent features \citep{hoyer2000independent,Gutmann14}. Deep networks behave similarly: the first-layer of CNNs trained for classification display achromatic and color-opponent units
\citep{krizhevsky2012imagenet,flachot2018processing}, and CNNs trained for low-level vision tasks show human-like color axes and behavior, including susceptibility to color illusions \citep{gomez2020color}, and distinctive spatial bandwidth for color channels~\citep{Li22}. Finally, vision foundation models, trained on natural images at a far larger scale, may therefore retain this color structure, and recent
evidence supports this expectation
\citep{akbarinia2025exploring,wickramanayaka2025systematic, gomez2025art}.

Most  modern text-to-image models are latent generators: rather than operating on pixels,
they generate within the representation learned by a VAE and decode the result back to image space \citep{rombach2022high}. VAEs are known to organize this representation around the factors of variation in the data: they tend to disentangle generative attributes \citep{higgins2017beta,burgess2018understanding} and to align their latent axes with the principal directions of the data \citep{rolinek2019variational,pmlr-v139-zietlow21a}. Yet, although color is one of the most thoroughly studied and highly structured properties of natural
images, how it is represented in these latent spaces has only recently begun to
be explored \citep{arias2025color,pach2026latent}. Meanwhile, text-to-image models still fail to bind colors to objects, render named colors faithfully, or follow numerical specifications such as HEX or RGB
codes \citep{Butt_2026_CVPR}. The color
structure may already be present in the latent space; but the generators simply do
not use it.

Recent studies have begun to probe color in these latent spaces with uniform images, recovering opponent hue axes in the Stable Diffusion VAE \citep{arias2025color} and an HSL-like subspace in the FLUX VAE \citep{pach2026latent}. Such inputs reveal where colors are mapped, but not how color is encoded alongside the spatial content of real images. We instead analyze the response to natural images and show that the VAEs of modern text-to-image models share a color subspace aligned with brightness and opponent-color directions (Fig.~\ref{fig:teaser}). On the encoder side, the singular vectors of a linear approximation of the encoder are achromatic and opponent-color patterns; on the decoder side, traversing individual latent channels shifts the color of the decoded image. We find this structure consistently in the VAEs of SD1.5, SDXL, SD3, FLUX, and FLUX.2, with 4 to 32 latent channels, and show that it is only partially disentangled from image structure. Building on it, we propose \method, which steers generation to follow numerical color specifications and achieves state-of-the-art accuracy on the CSS3/X11 system of GenColorBench (81.20\% vs.\ 51.96\%), and show that the same color basis also enables saturation control and color transfer.

In summary, our contributions are:
\begin{itemize}[leftmargin=10mm]
    \item We empirically show that the VAEs of modern text-to-image models share a color
    subspace aligned with brightness and nonlinear opponent-color axes, consistent across model families and latent sizes from 4 to 32 channels, and grounded on natural image statistics.
    \item We characterize this subspace through a linear approximation of the
    encoder and targeted latent steering, and quantify its partial entanglement
    with image structure.
    \item We propose a number of applications based on our characterization: \emph{\method} achieves state-of-the-art in precise
    numerical color generation on the fine-grained CSS3/X11 system of
    GenColorBench, \emph{saturation control} adjusts the global chromatic intensity, and \emph{color transfer} changes the palette to match a reference.
\end{itemize}

\section{Related Work}\label{sec:related_main}

\textbf{What VAEs disentangle.} Unsupervised disentanglement is impossible without inductive biases on both model and data \citep{locatello2019challenging}, yet VAEs often disentangle in practice \citep{higgins2017beta,burgess2018understanding}. \citet{rolinek2019variational} traced this to the diagonal posterior, which aligns the latent axes with the local principal directions of the data, and \citet{pmlr-v139-zietlow21a} showed that this yields disentanglement only when the variance structure of the data matches its generative factors. This line of work has been evaluated almost exclusively on synthetic benchmarks. We test it on color, whose natural-image statistics decorrelate into one luminance and two opponent axes \citep{buchsbaum1983trichromacy,ruderman1998statistics}.

\textbf{Color in vision foundation models.} The color representations of deep networks have been compared with human perception, with limited alignment \citep{wickramanayaka2025systematic,ehab2026vision}. For text-to-image VAEs, \citet{arias2025color} recovered opponent hue axes in the Stable Diffusion VAE and \citet{pach2026latent} an HSL-like subspace in the FLUX VAE, both by probing with uniformly colored patches. We instead analyze the response to natural images, across five VAEs with 4 to 32 latent channels.

\textbf{Precise color control.} Existing methods either act on the text conditioning, by learning color tokens \citep{butt2024colorpeel} or grounding HEX codes and RGB values in a CIELAB-aligned ColorBook (NumColor; \citealp{butt2026numcolor}), or intervene at inference time, either by rewiring cross-attention (ColorWave; \citealp{laria2026leveraging}) or by optimizing the latents \citep{shomer2026color}. Closest to ours, \citet{pach2026latent} recolor objects through an HSL mapping interpolated from eight anchor colors, but this approch is only specifical for FLUX. \method instead recovers the color basis of any pretrained VAE and learns the displacement toward the target from measured CIELAB outcomes. An extended discussion is given in Appendix~\ref{sec:related}.

\subsection{Background}
\label{sec:background}

\subsubsection{Variational autoencoders}
\label{subsec:VAEs}

Variational autoencoders (VAEs) \citep{kingma2014auto,rezende2014stochastic} learn a probabilistic mapping between images $\mathbf{x} \in \mathbb{R}^{d_x}$ and a lower-dimensional latent representation $z \in \mathbb{R}^{d_z}$. An encoder $E$ parameterizes an approximate posterior $q_\phi(z|\mathbf{x})$, typically a Gaussian with diagonal covariance, and a decoder $D$ parameterizes the likelihood $p_\theta(\mathbf{x}|z)$. Both are trained jointly by maximizing the Evidence Lower Bound (ELBO), which in its $\beta$-weighted form \citep{higgins2017beta} is defined by :
\begin{equation}
\mathcal{L}_{\beta} = \mathbb{E}_{q_\phi(z|\mathbf{x})}\left[\log p_\theta(\mathbf{x}|z)\right] - \beta\, D_{\text{KL}}\left(q_\phi(z|\mathbf{x}) \,\|\, p(z)\right),
\end{equation}
where the first term encourages faithful reconstruction and the second regularizes the posterior toward an isotropic Gaussian prior $p(z)=\mathcal{N}(\mathbf{0}, I)$ via the Kullback-Leibler divergence ($D_{\text{KL}}$). Setting $\beta > 1$ strengthens this regularization and is known to promote disentangled latent factors \citep{higgins2017beta,burgess2018understanding}.

The VAEs of modern latent generative models operate in the opposite regime. Following \citet{rombach2022high}, they are trained with $\beta \ll 1$ (on the order of $10^{-6}$), together with perceptual and adversarial losses \citep{esser2021taming}, which prioritizes reconstruction quality and keeps the latent space only mildly regularized. These VAEs are trained once and then kept frozen. The encoder compresses an image of size $H \times W \times 3$ into a spatial latent of size $\frac{H}{f} \times \frac{W}{f} \times C$, where $f$ is the spatial downsampling factor and $C$ is the number of latent channels, and the generator operates entirely in this latent space.

\subsubsection{Latent Diffusion and Flow Matching}
\label{subsec:generative}

Latent generative models learn to transform Gaussian noise into VAE latents $z_0 = E(\mathbf{x})$ through an iterative sampling process. Two formulations dominate current text-to-image models. Diffusion models \citep{ho2020denoising} define a forward process that progressively corrupts $z_0$ with Gaussian noise, $z_t = \sqrt{\bar{\alpha}_t}\, z_0 + \sqrt{1-\bar{\alpha}_t}\, \epsilon$ with $\epsilon \sim \mathcal{N}(\mathbf{0}, I)$, where  
$\bar{\alpha}_t$ decreases monotonically from 1 to 0 over time and train a network $\epsilon_\theta(z_t, t)$ to predict the added noise. Flow matching models \citep{lipman2023flow,liu2023flow}, adopted by SD3 and FLUX \citep{esser2024sd3}, instead connect data and noise through the linear path $z_t = (1-\sigma_t)\, z_0 + \sigma_t\, \epsilon$ , where $\sigma_t$ controls the noise magnitude, and train a network $v_\theta(z_t, t)$ to regress the velocity $\epsilon - z_0$ along it. Generation then reverses these processes, starting from pure noise and integrating toward a clean latent that is lated decoded by $D$ into an image.

In both formulations, the clean latent can be estimated in closed form at any intermediate step $t$ from the current network prediction, without completing the sampling trajectory \citep{song2021denoising}:
\begin{equation}
\hat{z}_0 = \frac{z_t - \sqrt{1-\bar{\alpha}_t}\, \epsilon_\theta(z_t, t)}{\sqrt{\bar{\alpha}_t}} \quad \text{(diffusion)}, \qquad
\hat{z}_0 = z_t - \sigma_t\, v_\theta(z_t, t) \quad \text{(flow matching)}.
\label{eq:x0_pred}
\end{equation}
Decoding $\hat{z}_0$ gives a provides an intermediate estimate of the final image, which we use to measure the color of the target object during generation (Sec.~\ref{sec:numerical}).

\section{Color in Autoencoder Latent Space}\label{sec:alignment}

We study how the latent spaces of autoencoders represent color in two steps. We first show empirically that color is directly readable from the raw latent channels: traversing a single channel moves the decoded image along a consistent color direction, within a limited range before its spatial content also changes (Sec.~\ref{sec_emp}). We then explain this behavior from the input side through a linear approximation of the encoder, whose singular vectors are opponent-colored patterns (Sec.~\ref{sec_lin_aproc}).

\subsection{Latent Channels Encode Color}\label{sec_emp}

In classical VAEs, individual latent coordinates often capture a single
factor of variation of the data: traversing one coordinate while keeping the
others fixed changes, for example, only the azimuth of a face
\citep{higgins2017beta,rolinek2019variational}. We apply this same probe, a
latent traversal, to the raw latent space of autoencoders, focusing on
color.

\begin{figure*}[t]
    \centering
    
    \begin{subfigure}[b]{0.49\textwidth}
        \centering        
        \includegraphics[width=1\textwidth]{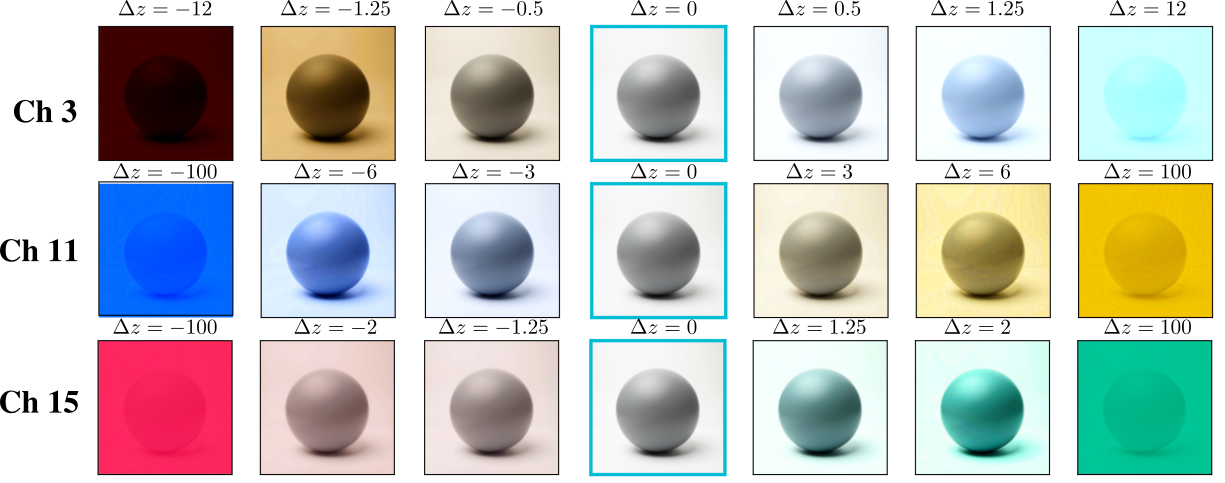}
        \caption{FLUX 2}
        \label{fig:dist_flux}
    \end{subfigure}
    \hfill
    \begin{subfigure}[b]{0.49\textwidth}
        \centering
        \includegraphics[width=1\textwidth]{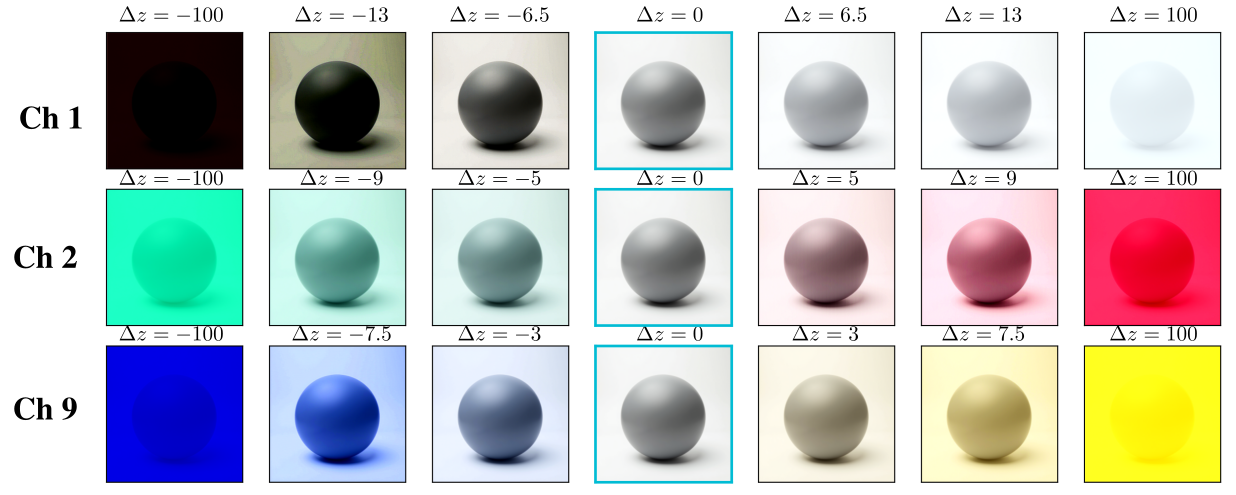}
        \caption{FLUX}
        \label{fig:dist_flux2}
    \end{subfigure}
    
    \vspace{0.15cm}

    \begin{subfigure}[b]{0.49\textwidth}
        \centering
        \includegraphics[width=1\textwidth]{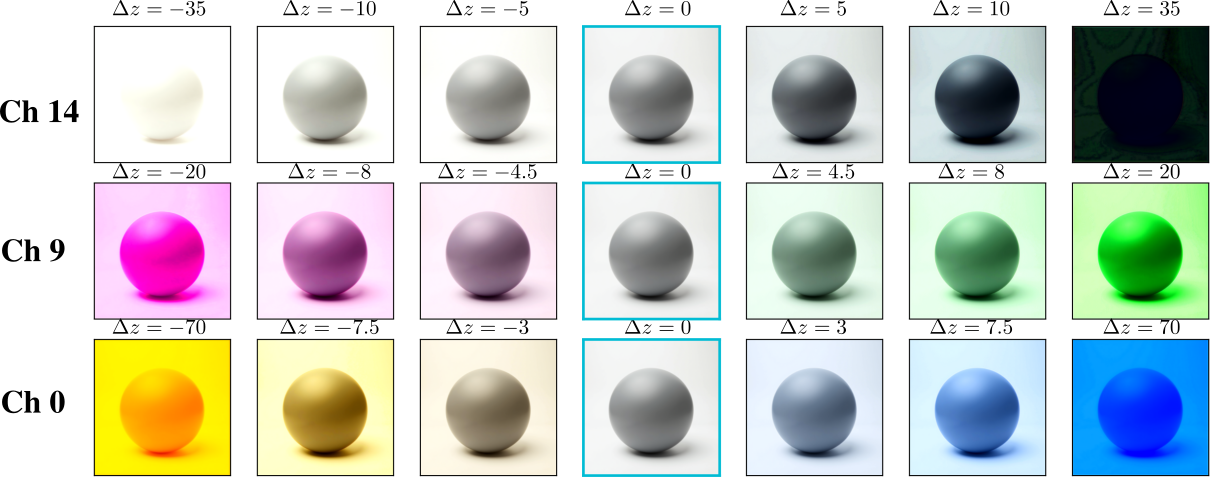}
         \caption{SD 3}
        \label{fi3g:eigen_sd3}
    \end{subfigure}
    \hfill
    \begin{subfigure}[b]{0.49\textwidth}
        \centering
        \includegraphics[width=1\textwidth]{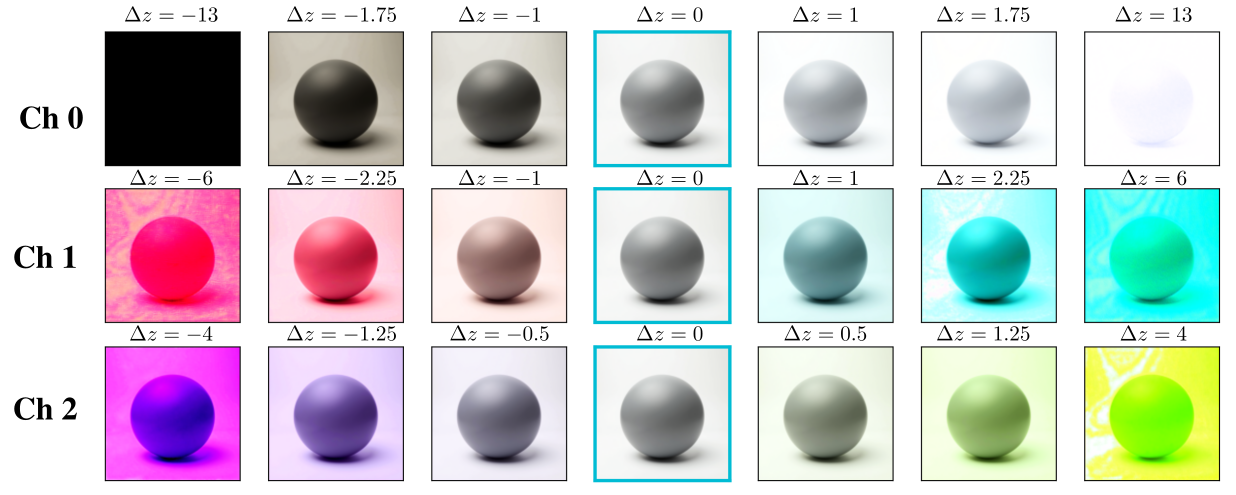}
        \caption{SDXL}
        \label{fig:dist_sdxl}
    \end{subfigure}
    
    \vspace{0.15cm}

    \vspace{-4mm}
    \caption{Image modifications through traverse channel variations across generative models: (a) Flux, (b) Flux 2, (c) Stable Diffusion 1.5, (d) SDXL, and (e) Stable Diffusion 3. }
    \label{fig:distortion}

\end{figure*}

Since these latents are spatial, we refer to each latent dimension as a
channel, and traverse channel $c$ by adding a constant offset $\delta$ at
every spatial position, $z_c \to z_c + \delta$, before decoding. As shown in
Fig.~\ref{fig:distortion}, modifying $\delta$ progressively shifts the
color of the decoded image along a consistent direction. We observe this
across deterministic autoencoders (AEs), VAEs, and VQ-VAEs with 3, 4, 8, and
32 channels trained from scratch, as well as in the pretrained
VAEs of SD1.5, SDXL, SD3, FLUX, and FLUX.2 (See the Fig.~\ref{fig:distortion} and Appendix \ref{sec:Latent_shift}). Color is therefore directly readable from the raw latent channels, without any learned probe.

\begin{figure}[h]
\begin{center}
\begin{subfigure}[b]{0.49\textwidth}
        \centering

        \includegraphics[width=1\textwidth]{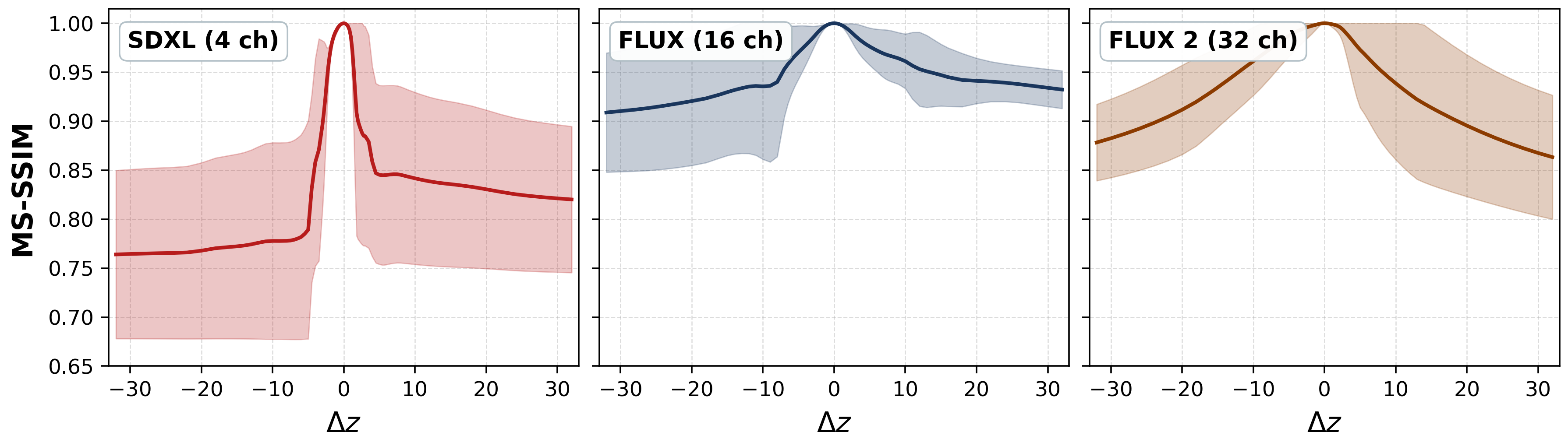}
        \caption{Texture distortion}
        \label{fig:ms-ssim}
    \end{subfigure}
\hfill
\begin{subfigure}[b]{0.49\textwidth}
        \centering
    
        \includegraphics[width=1\textwidth]{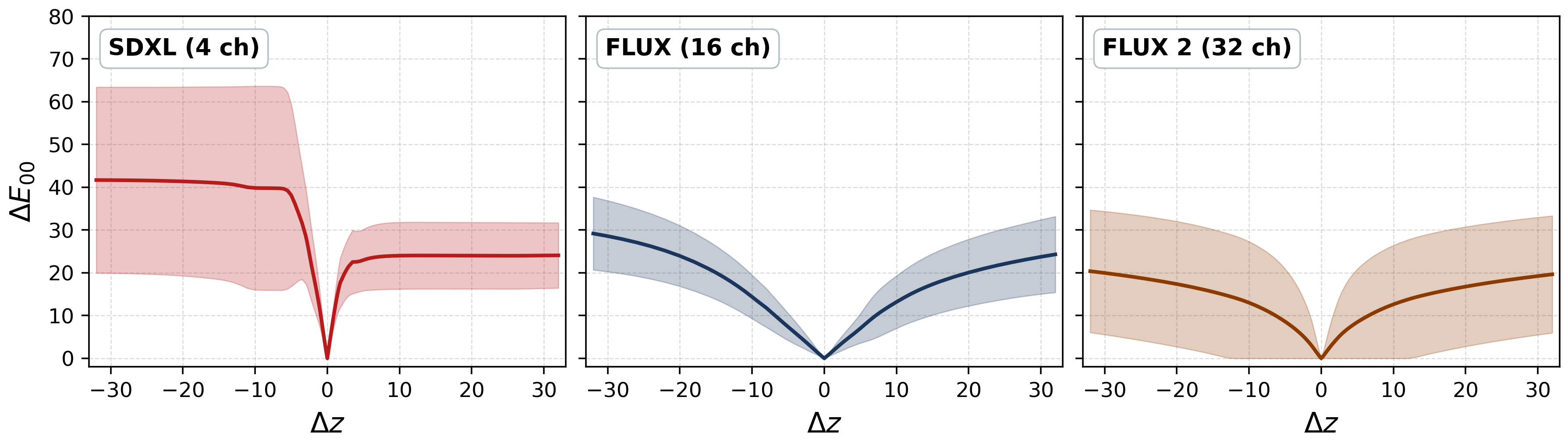}
        \caption{Color variation}
        \label{fig:deltae}
    \end{subfigure}

\end{center}
\vspace{-4mm}
\caption{Texture-color variations through traverse latent channel modifications in different VAEs.}
\vspace{-4mm}
\label{fig:delta_z}
\end{figure}

Color is, however, spatio-chromatically entangled: the same channels that
carry color also carry spatial information. For small offsets, traversing the channel modifies only the color of the decoded image; but beyond a certain magnitude, it also begins to alter its spatial content. This behavior is visible in the edges and textures in the outermost columns of Fig.~\ref{fig:distortion} (see Appendix \ref{sec:Latent_shift} for more examples). We quantify this in Fig.~\ref{fig:delta_z} as a function of $\delta$. We measure color change with $\Delta E_{00}$ (CIEDE2000) relative to the unperturbed reconstruction, and spatial change with MS-SSIM computed on grayscale images using only its contrast and structure terms, so that the intended color and brightness shifts are not penalized. As $|\delta|$ increases, $\Delta E_{00}$ grows steadily, while MS-SSIM stays close to 1 for small offsets and then
drops. Color can therefore be edited by moving along the latent channels,
but only within a limited range before the edit starts to change the spatial
content of the image.

In general, we find that individual channels do not uniquely map one-to-one onto specific colors: a single channel typically moves the image along a mixture of them, and several channels can move it in similar directions (e.g., two different channels may both turn the image green). A notable exception arises when there are only 3 latent channels. In this case, both the VAE and the AE
align closely with the opponent color space: one channel controls brightness,
and the other two control the opponent axes (see
Appendix \ref{sec:Latent_shift}). For the VAE, this is the behavior predicted by
\citet{rolinek2019variational}. For the AE, it is not: its objective has no mechanism that favors a particular orientation~\citep{rolinek2019variational}.
However, opponent color directions do appear in plain autoencoders~\citep{gomez2020color,Li22}, and, as shown below, they consistently emerge in modern autoencoders.

In the general case, where color is spread across correlated channels, the color subspace can still be recovered with a PCA decomposition of the channels' color responses, which yields one luminance and two opponent-color directions. This is the mechanism we exploit in our applications (Sec.~\ref{sec:aplic}).

\subsection{Linear Approximation of the VAE}\label{sec_lin_aproc}

The traversals in section \ref{sec_emp} show that color is readable from the latent channels, but not why. To answer this we turn to the input side and ask which image patterns the encoder is most sensitive to. Because a VAE is trained to reconstruct its input, its response stays close to the identity and is well described to first order, which lets us replace the encoder by a matrix whose singular vectors are precisely those patterns, ordered by how strongly they drive the latent. If the color structure observed above originates in the encoder, it should appear here as achromatic and opponent-color patterns among the leading singular vectors.

 Let $X \in \mathbb{R}^{d_x \times N}$ represent a matrix containing $N$ input images. Let $Z \in \mathbb{R}^{d_z \times N}$ represent their corresponding low-dimensional latent representations, obtained via the VAE's encoder, $Z = E(X)$. Subsequently, let $\hat{X} = D(Z)$ represent the reconstructed output images generated by the VAE's decoder. The application of the model (encoder + decoder) is denoted here as: $\mathbf{\hat{x}} = m(\mathbf{x})$. 

Following \citet{gomez2020color} and  \citet{Li22}, we analyze the behavior of the network through the linear approximation of its response. Considering any image $\mathbf{x}$ as a deviation from the average image (flat gray image, $\mathbf{0}$), the first-order approximation of the autoencoder is:
\begin{equation}
\mathbf{\hat{x}} = m(\mathbf{x}) \approx m(\mathbf{0}) + \nabla_{\mathbf{\mathbf{x}}} \, m(\mathbf{0}) \cdot \mathbf{x} = M_{vae} \cdot \mathbf{x}
\end{equation}
where the response to the average image is re-calibrated as zero, and $M_{vae} \in \mathbb{R}^{d_x \times d_x}$ is thus the Jacobian of the network response w.r.t. the input at 0. 
For VAE architectures, this assumption is justified by the ELBO objective, where the expected log-likelihood reconstruction term, $\mathbb{E}_{q_\phi(z|x)}[\log p_\theta(x|z)]$, forces the latent space to preserve sufficient information for data recovery. Furthermore, $M_{vae}$ can be decomposed as the product of the decoder and encoder matrices, such that $M_{vae} = M_{dec} M_{enc}$, where $M_{dec} \in \mathbb{R}^{d_x \times d_z}$ and $M_{enc} \in \mathbb{R}^{d_z \times d_x}$. 
We select the simplest method to compute $M_{enc}$ from the three approaches considered in \citet{Li22}. In particular, here $M_{enc} = Z \cdot X^\dag$, and $M_{dec} = \hat{X} \cdot Z^\dag$, where $^\dag$ denotes the pseudoinverse.

To analyze the structure of the latent space, we perform Singular Value Decomposition (SVD) on the encoder matrix, $M_{enc} = U \Sigma V^T$.

\begin{figure*}[t]
    \centering

        \centering
        
        \includegraphics[width=0.9\textwidth]{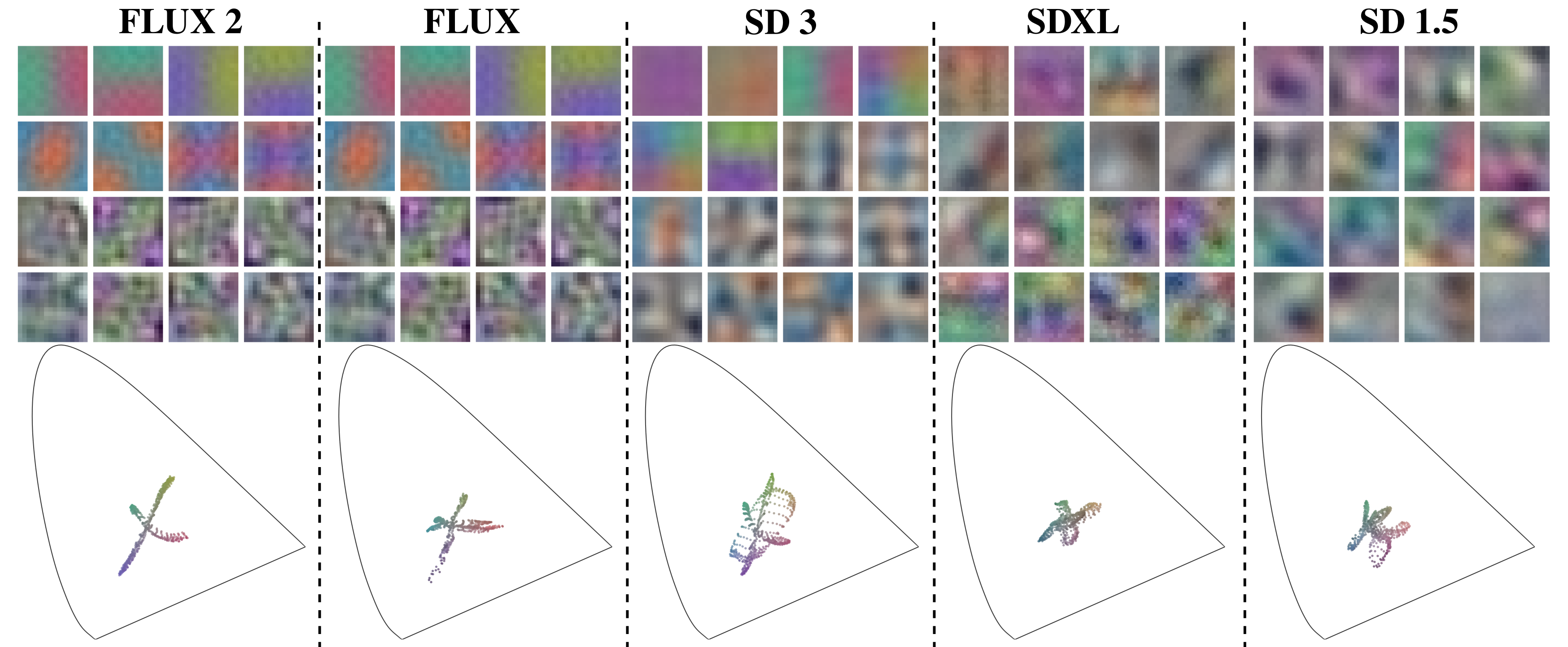}
         \caption{The first singular vectors of the linear approximation of VAEs and their corresponding CIE chromaticity directions across generative models: Flux.dev 2, Flux.dev 1, (c) SD 1.5, (d) SDXL, and SD 3. Each subplot visualizes the singular vectors ordered by their chromatic variance alongside their color content in the CIE chromatic diagram. We used $10^6$ $16\times16$ patches from \emph{Tiny ImageNet}. Appendix~\ref{larger} shows that this kind of vectors and opponent colors also emerge in other databases.}
         \vspace{-4mm}
    \label{fig:vae_eigenfunctions_grid}
  
\end{figure*}

In Fig. \ref{fig:vae_eigenfunctions_grid}, we visualize the singular vectors derived via SVD for the VAEs of text-to-image models. In addition, we plot the chromatic coordinates corresponding to the components with the largest singular values and highest chromatic variance on the CIE xy chromaticity diagram. Across all evaluated models, two principal opponent chromatic directions consistently emerge. This demonstrates that the latent spaces of these VAEs align with opponent color channels.

\textbf{Faithfulness of the linear approximation.} 
This reading of the encoder is only as informative as the linear approximation behind it. Appendix \ref{faith} shows that linear approximations of $E$ and $D$ are good descriptions of their behavior because (i) they explain a large fraction of the energy of the signal generated by the (nonlinear) autoencoder, and (ii) the errors wrt the original signal are small. 

\section{Applications}\label{sec:aplic}

Sec.~\ref{sec:alignment} shows that VAE latents carry a color structure inherited from natural image statistics, but spread across correlated channels and only partially disentangled from spatial content. We therefore recover an explicit color basis for each VAE, computed once (Sec.~\ref{sec:color_basis}), and use it for three applications in T2I generation, handling the residual entanglement at inference time: numerical color precision (Sec.~\ref{sec:numerical}), saturation control, and color transfer (Sec.~\ref{sec_more_apli}).

\subsection{Characterizing the Color Basis of a VAE}
\label{sec:color_basis}
Our goal is to find a small set of disentangled directions in the latent space.
Specifically, we seek directions that move the color of an image along the
luminance axis or the two opponent color axes independently. We extract these
principal color directions through the following steps.

\textbf{Step 1: Controlled perturbation.} We select seven reference objects spanning the primary axes of the CIELAB color space. For each object, we perturb one latent channel at a time. We add a scalar perturbation $z_{c,m} \to z_c + m$ inside the region of interest for a set of magnitudes $m$. We repeat this process for every channel $c$ in the latent space and measure how much the perceived color of the object shifts in the CIELAB space ($\Delta L^*_{c,m}, \Delta a^*_{c,m}, \Delta b^*_{c,m}$).

\textbf{Step 2: Measuring sensitivity.} For each channel, we compute the slope of color change per unit of perturbation:
\begin{equation}
s_{c,m,L} = \frac{\Delta L^*_{c,m}}{\Delta z_c}, \qquad s_{c,m,a} = \frac{\Delta a^*_{c,m}}{\Delta z_c}, \qquad s_{c,m,b} = \frac{\Delta b^*_{c,m}}{\Delta z_c}
\end{equation}
We then stack all values for different $m$ and  $c$ as rows to obtain a global sensitivity matrix $S \in \mathbb{R}^{C·m \times 3}$.

\textbf{Step 3: Isolating principal color directions.} We apply SVD directly to $S = U\Sigma V^\top$. The resulting three columns of $U \in \mathbb{R}^{(C \times m) \times 3}$ provide a basis that isolates the latent color shifts.

To use the principal color directions discovered for generation, we localize each perturbation to the target object using spatial masks from SAM 3~\citep{Sam3}. To prevent abrupt deviations from the natural generation path, we inject these perturbations only during the final generation steps, linearly decaying the steering weight to zero.

\subsection{Numerical Color Precision}\label{sec:numerical}
Numerical colour precision refers to the ability to generate an object with the exact RGB value or HEX code specified in the text prompt. State-of-the-art T2I models still struggle with this task, exhibiting a range of failure cases (see Fig.\ref{fig:errors_gen} and \cite{Butt_2026_CVPR, butt2026numcolor}). In this subsection, we present \method, which exploits the discovered directions to steer the generation towards the exact colour specified in the user prompt.

\begin{table}[t]
\centering
\caption{Numerical color accuracy (\%) on GenColorBench, on (a) the mini subset and (b) the full benchmark. The number of latent channels is shown in parentheses. \textbf{Bold}: best result for each backbone.}
\vspace{-3mm}
\label{tab:color_bench_metrics}
\footnotesize
\setlength{\tabcolsep}{4pt}

\sbox{\gcbfullbox}{%
\begin{tabular*}{0.485\linewidth}{@{\extracolsep{\fill}}lrrr@{}}
\toprule
Model & ISCC-L3 & CSS3/X11 & Overall \\
\midrule

SD3 (16)               &  5.92 &  4.63 &  5.06 \\
\quad + NumColor       & \textbf{39.46} & 47.17 & \textbf{44.32} \\
\quad + \method        & 12.63 & \textbf{54.74} & 39.02 \\
\addlinespace
SD3.5M (16)            &  7.53 &  5.78 &  6.36 \\
\quad + NumColor       & \textbf{43.35} & 46.87 & 45.57 \\
\quad + \method        & 40.65 & \textbf{75.79} & \textbf{62.65} \\
\addlinespace
FLUX (16)              &  7.74 &  5.86 &  6.49 \\
\quad + NumColor       & \textbf{48.02} & 51.96 & 50.50 \\
\quad + \method        & 43.15 & \textbf{81.20} & \textbf{64.13} \\
\bottomrule
\end{tabular*}}%
\begin{subtable}[t]{0.485\linewidth}
\centering
\caption{GenColorBench-mini}
\label{tab:gcb_mini}
\parbox[t][\dimexpr\ht\gcbfullbox+\dp\gcbfullbox\relax][c]{\linewidth}{%
\begin{tabular*}{\linewidth}{@{\extracolsep{\fill}}lrrr@{}}
\toprule
Model & ISCC-L3 & CSS3/X11 & Overall \\
\midrule
SDXL (4)               &  1.84 &  7.37 &  5.53 \\
\quad + ColorWave      & 11.15 & 28.05 & 22.42 \\
\quad + \method        & \textbf{30.78} & \textbf{63.97} & \textbf{52.90} \\
\addlinespace
Z-Image (16)           & 16.92 & 12.45 & 13.94 \\
\quad + \method        & \textbf{44.91} & \textbf{81.11} & \textbf{59.22} \\
\addlinespace
FLUX.2 (32)            & \textbf{39.29} & 42.67 & 41.54 \\
\quad + \method        & 38.91 & \textbf{81.90} & \textbf{64.57} \\
\bottomrule
\end{tabular*}}
\end{subtable}%
\hfill
\begin{subtable}[t]{0.485\linewidth}
\centering
\caption{GenColorBench (full)}
\label{tab:gcb_full}
\parbox[t][\dimexpr\ht\gcbfullbox+\dp\gcbfullbox\relax][c]{\linewidth}{\usebox{\gcbfullbox}}
\end{subtable}
\end{table}

\begin{figure}[t]
    \centering
    \includegraphics[width=0.99\textwidth, height=0.85\textheight, keepaspectratio]{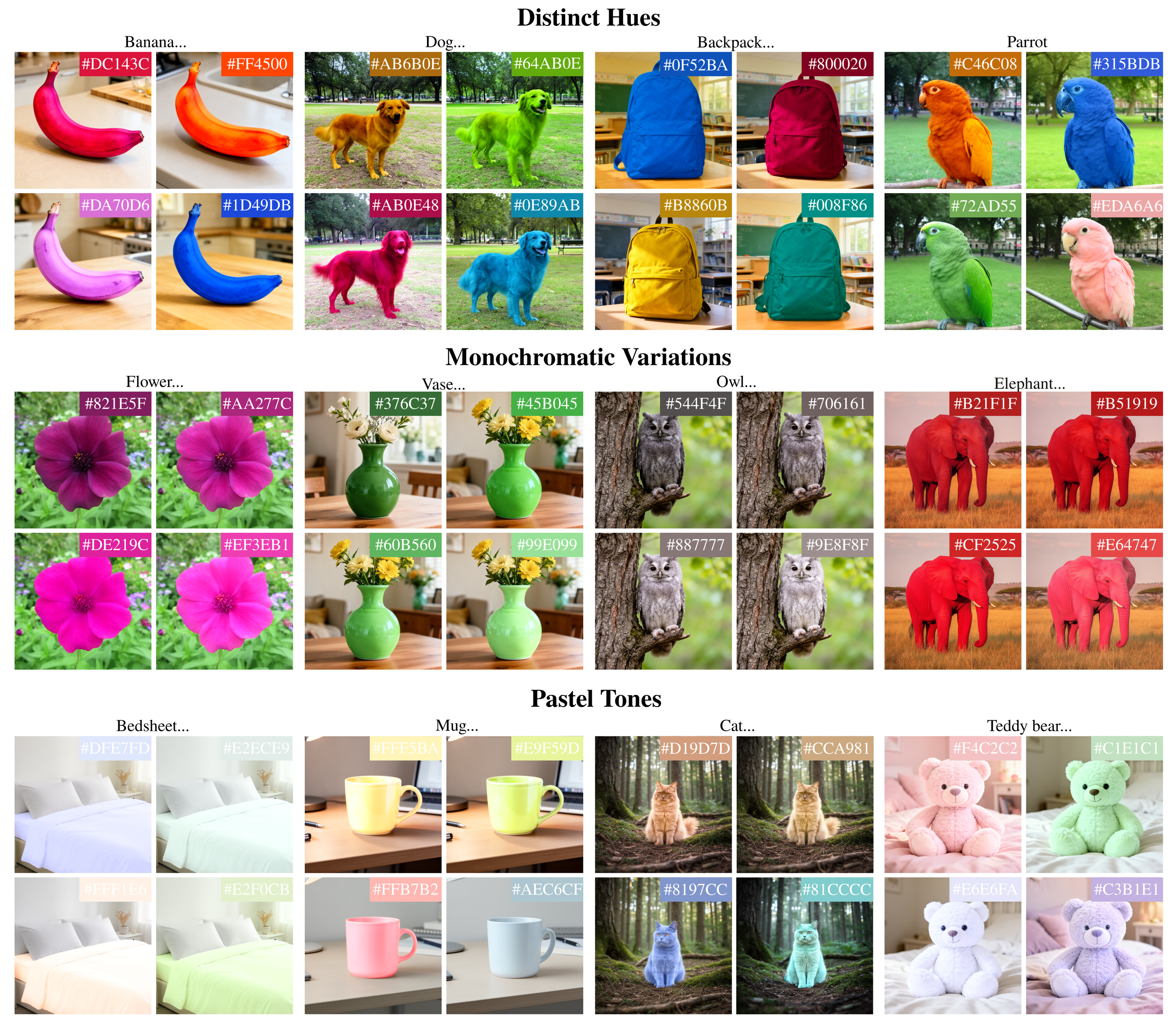}
    \vspace{-3mm}
    \caption{Qualitative results of \method. The generated examples demonstrate precise color control across three scenarios: applying distinct hues (top), generating subtle monochromatic variations within the same color family (middle), and pastel tones (bottom).}
    \vspace{-4mm}
    \label{fig:qualitative_res}
\end{figure}

\subsubsection{ Steering the Generator}

Our approach is based on two key properties of the discovered color directions. First, diffusion and flow-matching
models resolve low spatial frequencies first \cite{biroli2024dynamical}: the coarse color of a region is
settled well before sampling ends, so decoding the predicted clean latent
$\hat{z}_0$ (Eq.~\ref{eq:x0_pred} in Appendix~\ref{sec:background}) at an intermediate step already reveals the
color the object will have in the final image. Second, the characterization of
Sec.~\ref{sec:color_basis} gives the directions along which that color can be
moved. Measuring the current color and applying the corresponding displacement
over the remaining steps is therefore enough to reach the target (see Fig.~\ref{fig:method} in Appendix~\ref{app:pipelines}).

\textbf{Training.}
The basis of Sec.~\ref{sec:color_basis} fixes the directions, but not the
magnitude: how far a given displacement moves the object in CIELAB depends on
the color it starts from, since the decoder is nonlinear. We therefore learn
this relation empirically. For a set of prompts containing a color name, we generate one image without intervention and, for each displacement $m \in \mathcal{M}$ along the basis, a second image with the perturbation applied to the target object. Measuring the color of the object in both images gives a triplet: the color before the perturbation, the color it reached, and the displacement that produced it. On this data we train a small MLP that takes an initial and a target color, $C_i = (L_i, a_i, b_i)$ and $C_t = (L_t, a_t, b_t)$, and predicts the displacement $(m_1, m_2, m_3)$ along the basis that carries $C_i$ to $C_t$ (Alg.~\ref{alg:training}).

\textbf{Inference.}
The prompt still has to place the object in the scene, so we never pass the
numerical code to the text encoder. We replace it with a proxy color name: the
nearest entry in ISCC-NBS Level 2 to the requested value, measured in CIELAB. A prompt such as \emph{``a car of color \colorbox[HTML]{7B3F00}{\textcolor{white}{\#7B3F00}}''} becomes \emph{``a \colorbox[HTML]{7B3F00}{\textcolor{white}{brown}} car''}. The generator renders names far more reliably than codes, so the trajectory starts in the neighborhood of the target color and the steering only has to cover the residual difference. At a gate step $s$, we predict the clean latent $\hat{z}_0$ (Eq.~\ref{eq:x0_pred}), decode it, locate the object with SAM~3 and measure its mean color in CIELAB. The MLP maps this color and the target to a displacement along the color basis, which we add to the latent inside the object mask at each remaining step, with a weight that decays linearly to zero so that the final steps re-integrate the perturbation into the image structure (Alg.~\ref{alg:inference}; hyperparameters in Appendix~\ref{app:pipelines}).

\subsubsection{Results of \method}
\label{sec:numerical_results}

\textbf{Setup.}
We evaluate \method on GenColorBench~\citep{Butt_2026_CVPR}, a benchmark for precise color generation in which prompts specify object colors as HEX codes and RGB values. We report accuracy on its two color systems, ISCC-L3 and CSS3/X11, under both formats. We apply \method\ to six generators covering the three latent sizes analyzed in Sec.~\ref{sec:alignment}: SDXL (4 channels), SD3, SD3.5M, FLUX and Z-Image (16), and FLUX.2 (32). We compare against the vanilla models and against two color-control methods, each on the backbones for which it is available: ColorWave on SDXL and NumColor on SD3, SD3.5M and FLUX.

\textbf{Quantitative results.}
On the full benchmark (Table~\ref{tab:gcb_full}), vanilla models stay below 8\% on both systems. On CSS3/X11, NumColor reaches 47--52\%, while \method achieves the best accuracy on every backbone, up to 81.20\% on FLUX. This is remarkable because, while NumColor remains ahead on ISCC-L3, the CSS3/X11 on which \method dominates is the one on which every model evaluated in GenColorBench performs worst~\citep{Butt_2026_CVPR}. On the mini subset (Table~\ref{tab:gcb_mini}), \method brings FLUX.2 and Z-Image to about 81\% on CSS3/X11 and more than doubles ColorWave on SDXL (63.97\% vs.\ 28.05\%). A 2AFC user study with 22 observers with normal color vision confirms these gains perceptually: \method is preferred over both NumColor (FLUX) and ColorWave (SDXL), with z-scores (computed using Thurstone’s model) of $2.62 \pm 0.30$ and $1.67 \pm 0.30$, respectively (95\% CI; 0 indicates no preference). Details are in Appendix~\ref{app:user_study}.

\textbf{Qualitative results.}
Fig.~\ref{fig:qualitative_res} presents qualitative results obtained by applying \method to FLUX.2. The first two rows illustrate the method's behavior across the full color spectrum, showing that the same concept can be rendered in substantially different hues while preserving its structure and appearance. The third and fourth rows evaluate fine-grained numerical color control, where target colors differ only subtly. The final two rows focus on pastel colors, which combine high luminance with low saturation and are particularly challenging for contemporary generative models. Even in this demanding setting, \method faithfully reproduces the requested colors.

\subsection{Further Color Applications During Generation}\label{sec_more_apli}

\begin{figure}[t]
    \centering
    \begin{subfigure}[b]{\linewidth}
        \centering
        \includegraphics[width=0.94\linewidth]{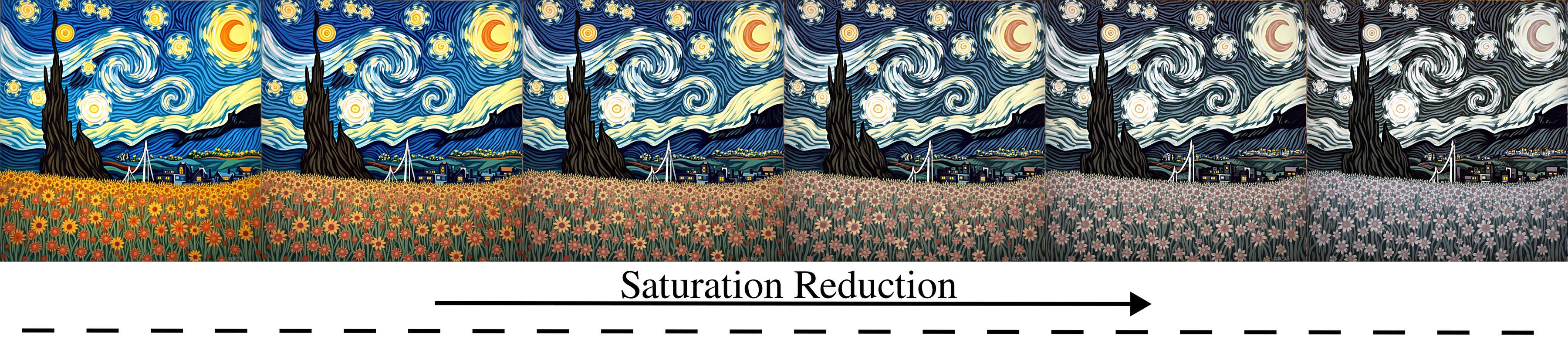}
        
        \label{fig:sat}
    \end{subfigure}

    \vspace{0.2cm}

    \begin{subfigure}[b]{\linewidth}
        \centering
        \includegraphics[width=0.90\linewidth]{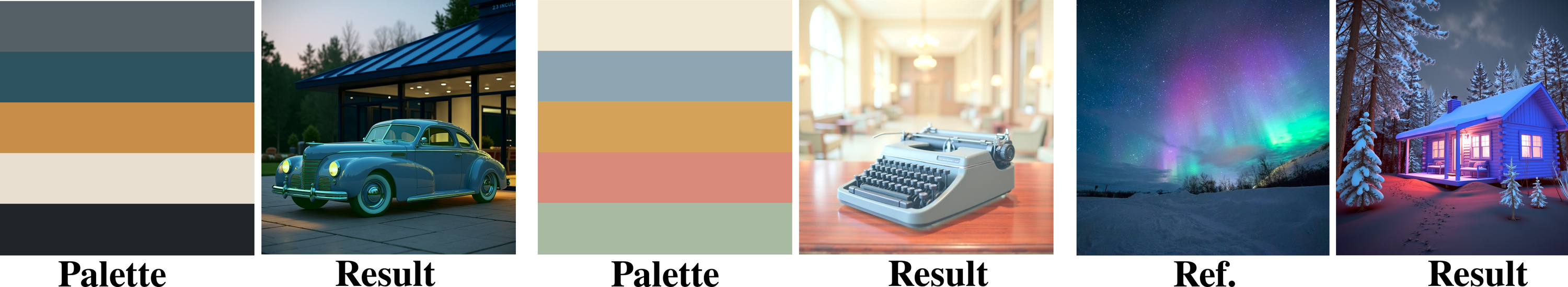}
       
        \label{fig:transfer}
    \end{subfigure}
    \caption{Further color control with the same color basis. \textbf{Top:} saturation reduction. From left to right, we intervene during generation to reduce the saturation of the final image. \textbf{Bottom:} color transfer, three examples. In each pair, the left image is the color reference and the right image is the generated result, whose colors are steered to match it.}
    \vspace{-3mm}

    \label{fig:further_apps}
\end{figure}

The same characterization naturally extends to other forms of color control at generation time, without requiring any modification to the procedure introduced in Sec.~\ref{sec:color_basis}. Fig.~\ref{fig:further_apps} (top) illustrates saturation control: by traversing the corresponding latent directions, our method can continuously modulate the saturation of the generated image (in the figure from $100\%$ to $50\%$ at $10\%$ intervals). Such saturation-aware generation could facilitate the synthesis of images within narrower color gamuts, potentially reducing the need for post-processing gamut mapping techniques~\citep{morovic2008gamut}.

Fig.~\ref{fig:further_apps} (bottom) shows a second application: color transfer during generation. In this setting, the input consists of both a text prompt and a reference image whose color distribution should be reproduced. Our method leverages the discovered latent directions to match the color distribution of the reference image while simultaneously generating the content specified by the prompt. Full details of both applications are given in Appendix~\ref{app:further}.

\section{Conclusions}

We have shown that color structure is already present in the raw latents of autoencoders trained on natural images: in AEs, VAEs, and VQ-VAEs alike, traversing a single latent channel shifts the color of the decoded image. This observation can be explained by a linear approximation of the encoder, whose leading singular vectors correspond to the achromatic and opponent-color patterns predicted by natural image statistics. Recovering these axes as an explicit basis turns the observation into a tool for color control during generation. Our main use of it is \method, which achieves state-of-the-art accuracy in numerical color generation on the fine-grained CSS3/X11 system of GenColorBench; saturation control and color transfer further show, qualitatively, that the same basis supports other forms of color control.

\paragraph{Limitations.} Existing methods for encouraging precise color synthesis in generative models typically rely on learning new concepts that enforce specific color relationships. In contrast, our approach leverages the inherent color alignment already present in the VAE latent space to dynamically guide the generation process. However, this strategy remains sensitive to large steering magnitudes. As shown in our analysis, the extracted latent color channels are not fully disentangled from spatial and structural features. Consequently, when the MLP predicts a large displacement from the source color toward the target one, the resulting latent perturbation affects not only color but also image content, leading to structural distortions in the final output.

\paragraph{Acknowledgments}
This work was supported by Grant, PID2024-162555OB-I00 funded by MCIN/AEI/10.13039/ 501100011033 and by ERDF "A way of making Europe", AIA2025-163919-C52 funded by MCIN/AEI/10.13039/501100011033, the grant Catedra ENIA UAB-Cruïlla (TSI-100929-2023-2) from the Ministry of Economic Affairs and Digital Transformation of Spain, the Generalitat de Catalunya CERCA Program, and partially funded by the Spanish Government and the EU under the MCIN/AEI/FEDER/UE Grant PID2023-152133NB-I00, as well as by the BBVA Foundations of Science program in Maths, Stats, Comp. Sci. and AI, grant VIS4NN: Vision Science for Artificial Neural Networks. We acknowledge the EuroHPC Joint Undertaking for awarding us access to Leonardo at CINECA, Italy, under project EHPC-AIF-2026SC01-004. JVC also acknowledges the Generalitat de Catalunya CERCA Program, and the 2025 Leonardo Grant for Scientific Research and Cultural Creation from the BBVA Foundation. The BBVA Foundation accepts no responsibility for the opinions, statements and contents included in the project and/or the results thereof, which are entirely the responsibility of the authors. Kai Wang acknowledges funding No.R002026G0116 and No.R002026B0121 from Guangdong and Hong Kong Universities 1+1+1 Joint Research Collaboration Scheme, No.TIC2026SP005 from Seed Project under City University of Hong Kong (Dongguan) Technology and Innovation Center, NSFC Grant No. 62606428, funding No.2025QN12X071 from Guangdong Pearl River Talent Program and the start-up grant B01040000108 from CityU-DG.

\newpage
\bibliography{iclr2027_conference}
\bibliographystyle{iclr2027_conference}

\newpage
\appendix
\section{Appendices}

\subsection{LLM usage statement}
LLMs were used for grammar correction and to rewrite and refine parts of the manuscript. They also assisted with code development.

\subsection{Related Work}\label{sec:related}

\subsubsection{Representation Learning and Disentanglement in VAEs}

Whether a VAE recovers the true factors of variation in its data is a question of identifiability. \citet{locatello2019challenging} showed that unsupervised disentanglement is impossible without inductive biases on both model and data, and identifiability has since been proven only under additional assumptions such as auxiliary variables \citep{khemakhem2020variational}. Yet VAEs often disentangle in practice \citep{higgins2017beta,burgess2018understanding}. \citet{rolinek2019variational} traced this to the diagonal posterior, which makes the decoder Jacobian locally orthogonal and aligns the latent axes with the local principal directions of the data. Relatedly, \citet{reizinger2022embrace} connected the ELBO to independent mechanism analysis in the near-deterministic regime. \citet{pmlr-v139-zietlow21a} further showed that this bias yields disentanglement only when the variance structure of the dataset matches its generative factors. What a VAE disentangles is therefore dictated by the principal directions of its training data.

Color is a natural test of this view: it is one of the most structured properties of natural images, whose statistics decorrelate into one luminance and two opponent-color axes \citep{buchsbaum1983trichromacy,ruderman1998statistics}. Yet this line of work has been evaluated almost exclusively on synthetic benchmarks, where color appears at most as a discrete hue label, and has not examined how VAEs trained on natural images represent it. Only recently have studies begun to probe color in the latent spaces of text-to-image VAEs \citep{arias2025color,pach2026latent}, and none relates the structure they find to the statistics of the training data. We instead analyze the encoder's response to natural images and show that, across model families and latent sizes, these latents share a color subspace aligned with the luminance and opponent axes that natural image statistics would predict.

\subsubsection{Color Representations in Vision Foundation Models}

Recent work has begun to characterize how color is encoded in the representations of vision foundation models. \citet{wickramanayaka2025systematic} compared the internal color representations of 16 deep networks with human similarity judgments over 93 colors, and found that only CLIP models reached fine-grained structural agreement with human perception. \citet{ehab2026vision} probed more than 50 vision encoders, including the SD3 VAE, and compared their chromatic discrimination thresholds with human discrimination ellipses, finding weak alignment across all model families. For generative models, \citet{arias2025color} encoded uniformly colored patches with the Stable Diffusion VAE and applied PCA to their spatially averaged latents, recovering an intensity axis and two opponent hue axes. When probing grayscale synthetic patterns separately, they found that one of these chromatic channels also carries shape information. More recently, \citet{pach2026latent} identified an HSL-like color subspace in the VAEs of FLUX, also by probing them with uniformly colored images, and used it to steer the color of generated objects. These studies characterize color through flat stimuli, which reveal where colors are mapped but not how color is encoded alongside the spatial content of real images. Moreover, each examines only one or two VAEs, leaving open whether the structure they find is specific to a particular model or shared across text-to-image VAEs with different architectures and latent sizes.

\subsubsection{Precise Color Control in Text-based Image Models}

Methods for precise color control in text-to-image models intervene at two points of the generative pipeline. A first family acts on the text conditioning. ColorPeel \citep{butt2024colorpeel} learns color prompt tokens from synthetic stimuli that disentangle color from shape, but covers only a few colors per training session and is only able to interpolate between nearby colors. Targeting numerical specifications, NumColor \citep{butt2026numcolor} addresses the fragmentation of hex and RGB codes into meaningless subword tokens by aggregating them into a single unit and grounding it with a ColorBook of learnable embeddings aligned with CIELAB. 

A second family intervenes at inference without retraining the backbone, either by rewiring the cross-attention of IP-Adapter to bind a color descriptor to a patch of the target color~\citep{laria2026leveraging}, or by optimizing the latents with attention and color losses~\citep{shomer2026color}. Closest to our work, \citet{pach2026latent} recolor objects by shifting latents within the color subspace discussed above, through a mapping to HSL interpolated from eight anchor colors. Their interpretation is derived for FLUX alone, from flat color patches. We instead build on a general property: the latents of VAEs trained on natural images align with luminance and opponent-color axes, a structure inherited from natural image statistics (Sec.~\ref{sec:alignment}). This lets us design a characterization procedure that recovers these directions in any pretrained VAE, along which \method learns the displacement toward the target from measured CIELAB outcomes rather than from a fixed geometric mapping. 

\subsection{ Latent Channel Shifts}\label{sec:Latent_shift}
\subsubsection{Custom Baselines}
We trained three autoencoder families from scratch on Tiny ImageNet-200~\citep{le2015tiny} at its native $64 \times 64$ resolution:
  deterministic autoencoders (\textbf{AE}), continuous variational autoencoders (\textbf{KL-VAE}), and vector-quantized autoencoders
  (\textbf{VQ-VAE}). All models share a unified convolutional backbone~\citep{rombach2022high} with a spatial downsampling factor of $f=4$ and were evaluated across bottleneck widths $C \in \{3, 4, 8, 16\}$. Training was performed for $40$
  epochs using Adam ($\beta_1 = 0.5, \beta_2 = 0.9$) with a learning rate of $4.5 \times 10^{-6}$.

    To examine how chromatic information is encoded across the latent space, we performed latent shift interventions on encoded grayscale
  test images. The resulting channel-wise response grids are shown in Figure~\ref{fig:ae_channel_shifts} for the AE, Figure~\ref{fig:kl_channel_shifts} for KL-VAE, and Figure~\ref{fig:vq_channel_shifts} for VQ-VAE.
\begin{figure}[htbp]
    \centering
   
    \begin{subfigure}[c]{0.48\textwidth}
        \centering
        \includegraphics[width=\textwidth]{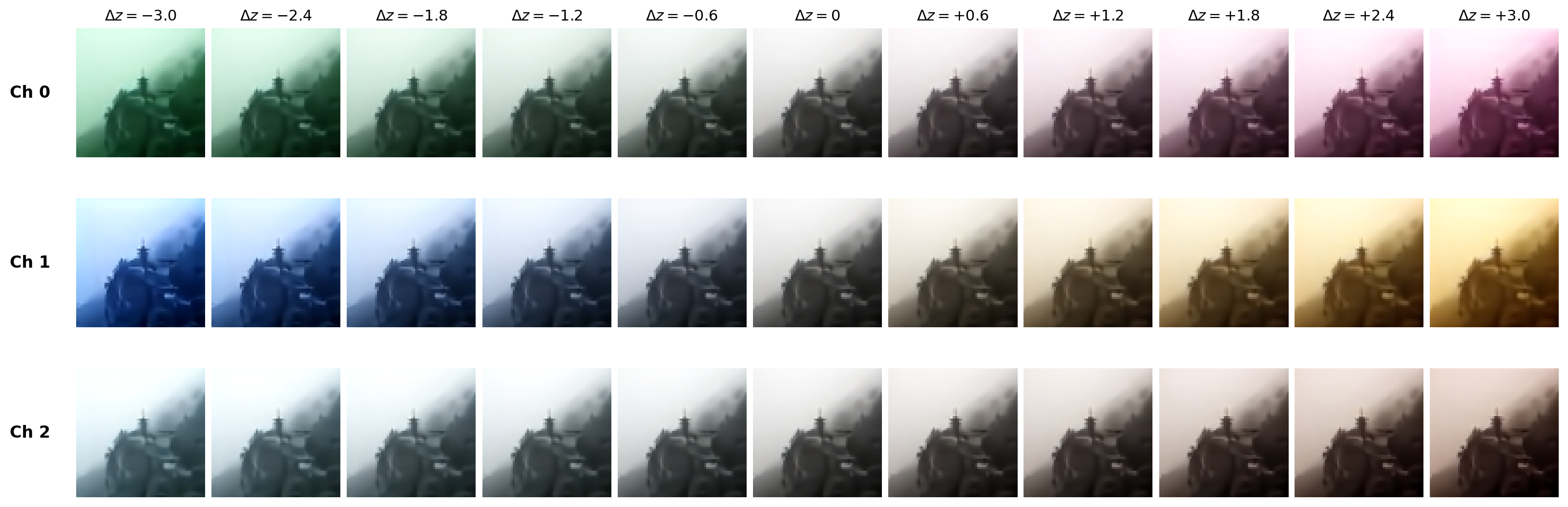}
        \caption{3-channel AE}
        \label{fig:shift_3ch}
    \end{subfigure}
    \hfill
    \begin{subfigure}[c]{0.48\textwidth}
        \centering
        \includegraphics[width=\textwidth]{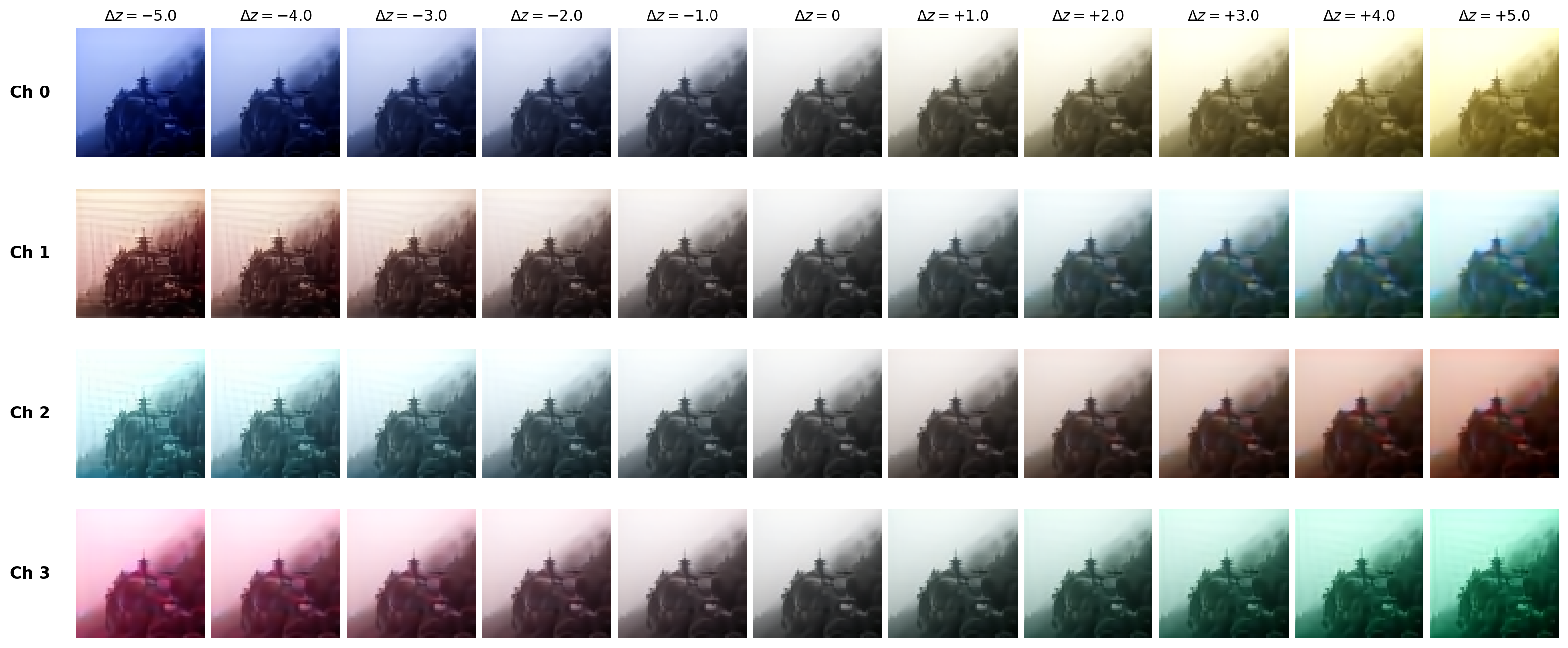}
        \caption{4-channel AE}
        \label{fig:shift_4ch}
    \end{subfigure}
    
    \vspace{0.4cm}

    \begin{subfigure}[c]{0.48\textwidth}
        \centering
        \includegraphics[width=\textwidth]{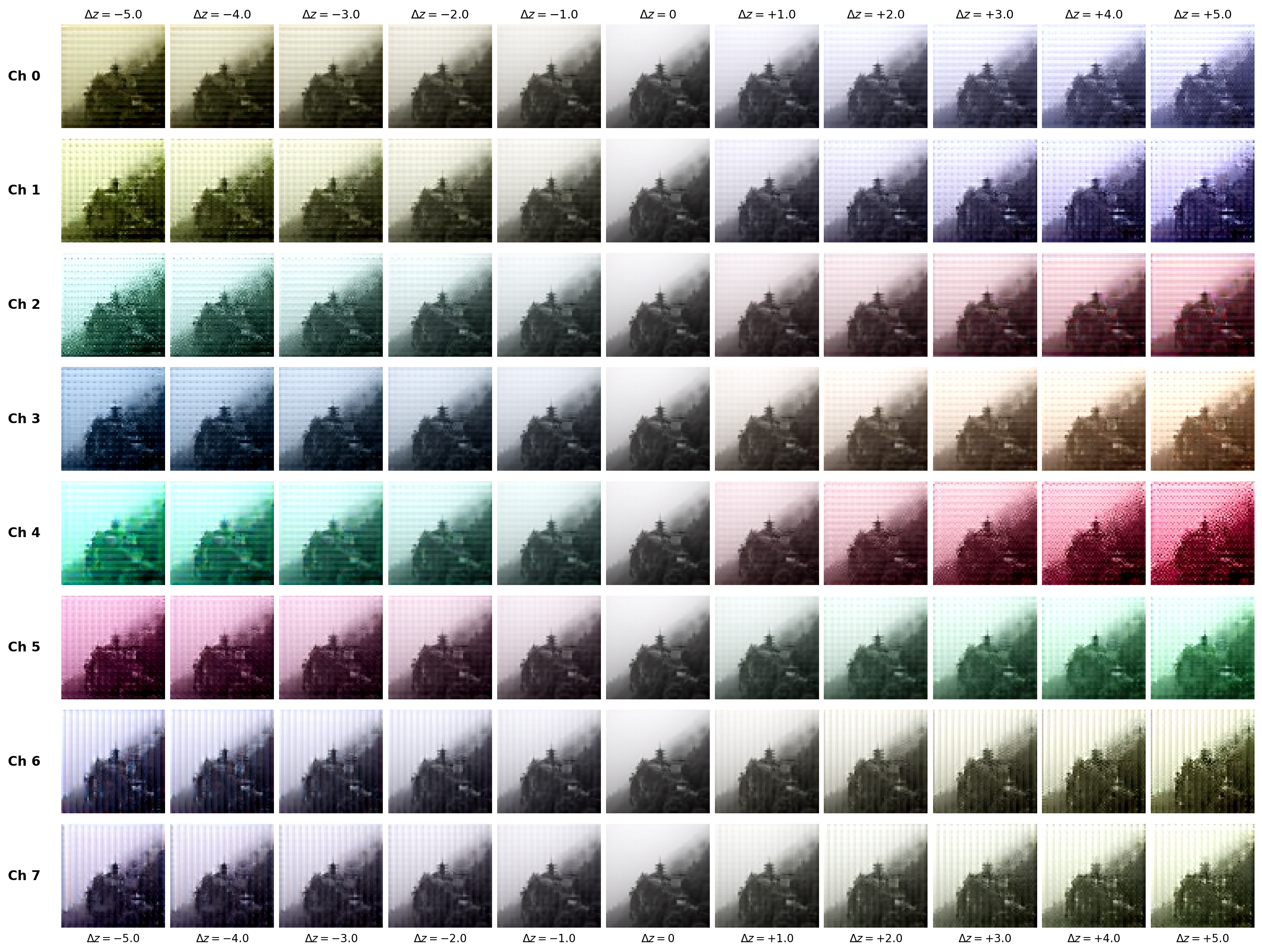}
        \caption{8-channel AE}
        \label{fig:shift_8ch}
    \end{subfigure}
    \hfill
    \begin{subfigure}[c]{0.48\textwidth}
        \centering
        \includegraphics[width=\textwidth]{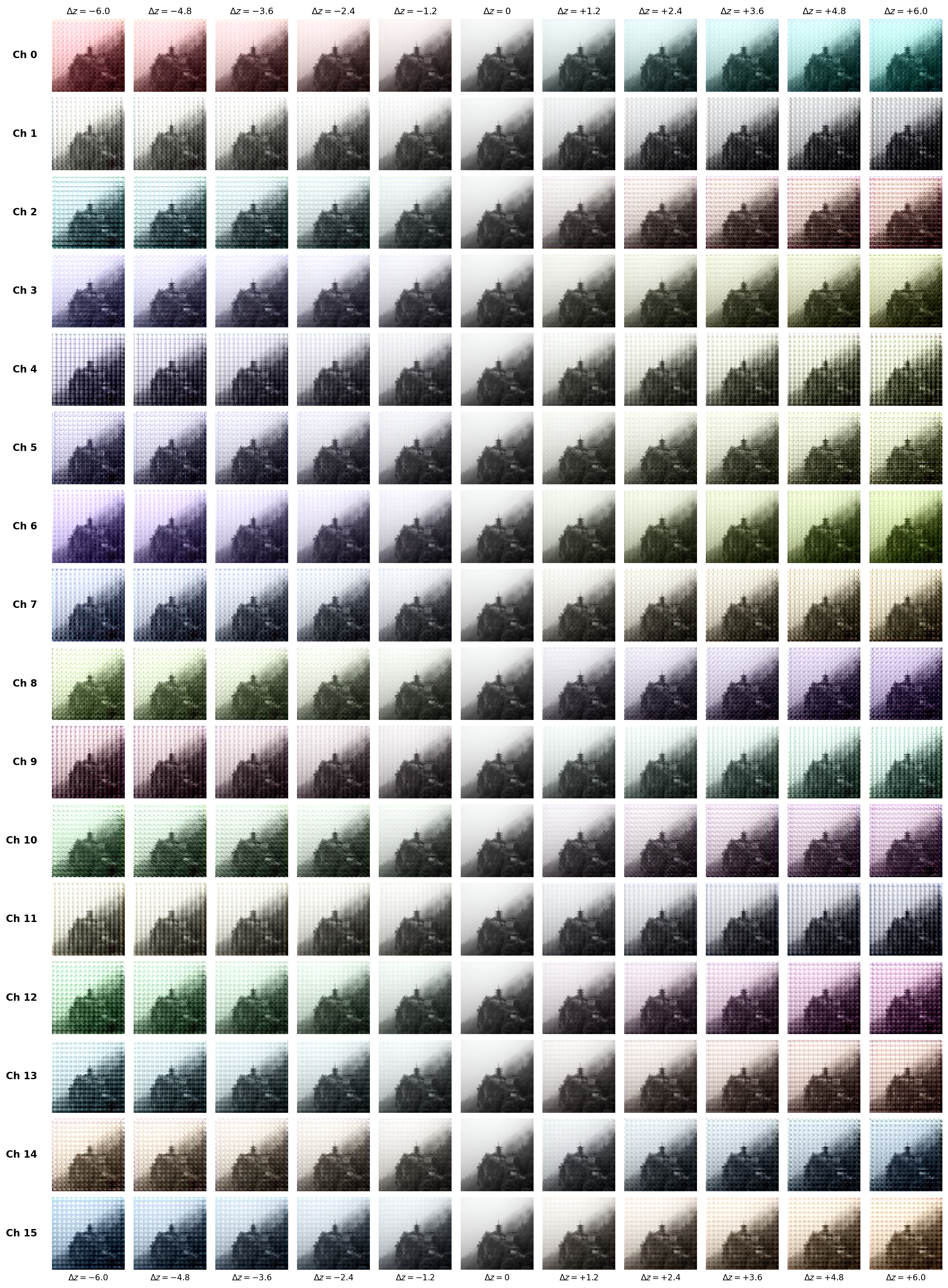}
        \caption{16-channel AE}
        \label{fig:shift_16ch}
    \end{subfigure}
    
    \caption{Visual impact of individual latent channel shifts ($\Delta z$) across different AE.}
    \label{fig:ae_channel_shifts}
\end{figure}

\begin{figure}[htbp]
    \centering

    \begin{subfigure}[c]{0.48\textwidth}
        \centering
        \includegraphics[width=\textwidth]{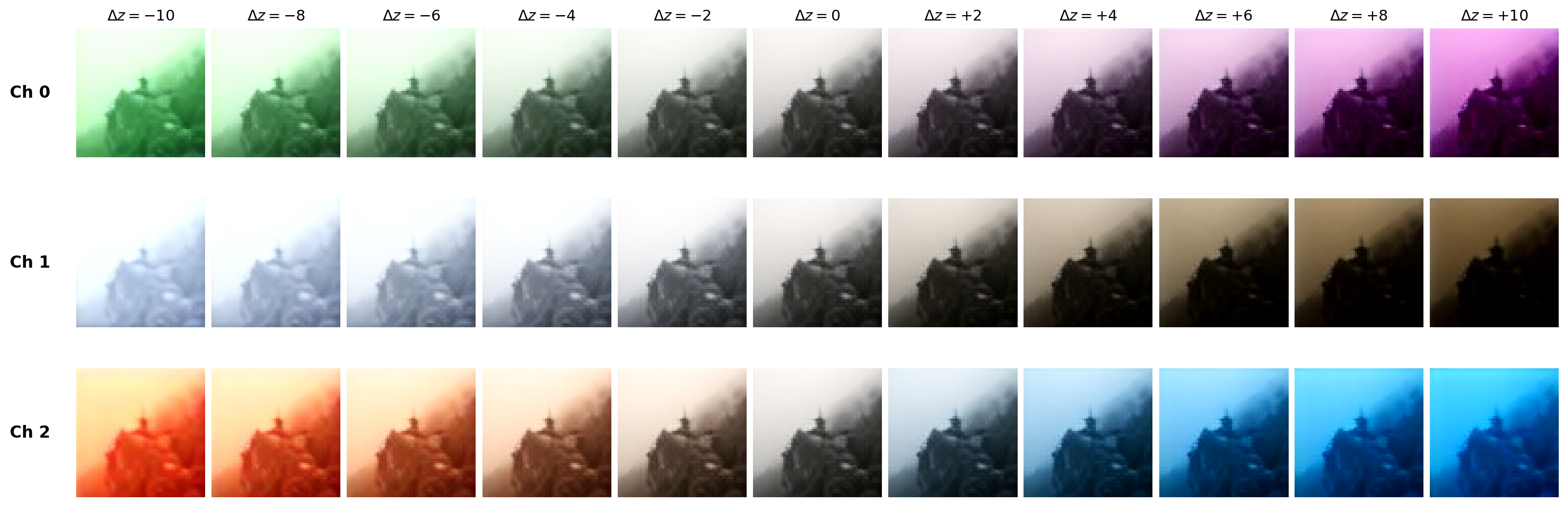}
        \caption{3-channel KL-VAE}
        \label{fig:shift_3ch}
    \end{subfigure}
    \hfill
    \begin{subfigure}[c]{0.48\textwidth}
        \centering
        \includegraphics[width=\textwidth]{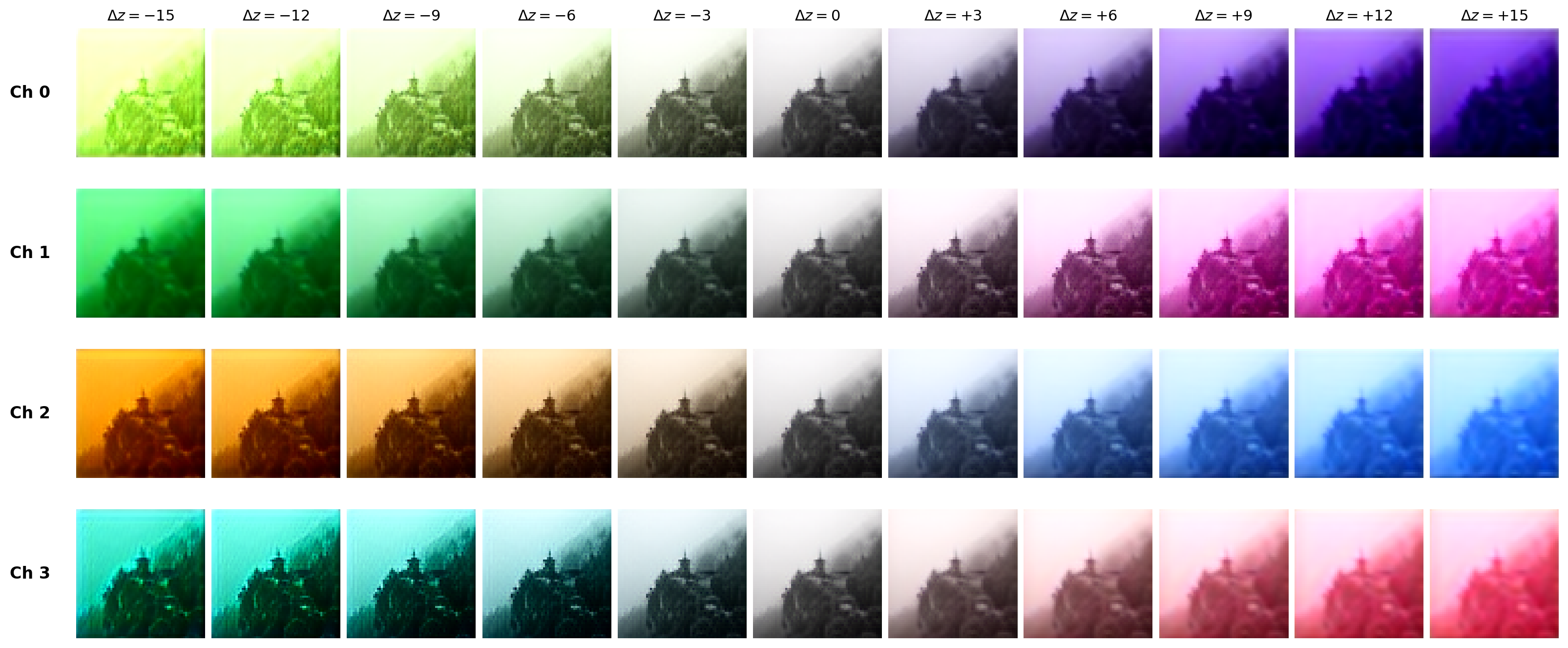}
        \caption{4-channel KL-VAE}
        \label{fig:shift_4ch}
    \end{subfigure}
    
    \vspace{0.4cm}

    \begin{subfigure}[c]{0.48\textwidth}
        \centering
        \includegraphics[width=\textwidth]{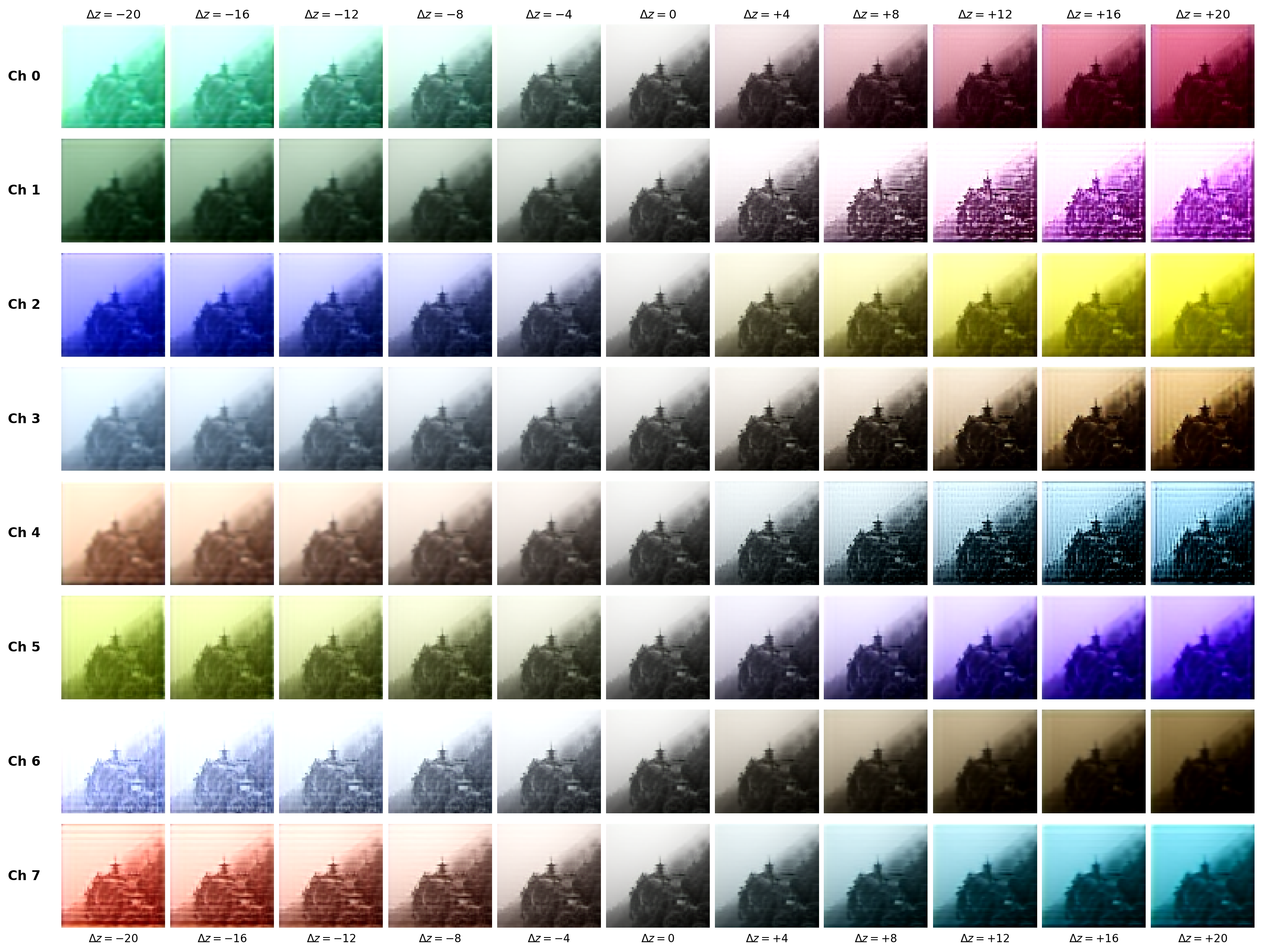}
        \caption{8-channel KL-VAE}
        \label{fig:shift_8ch}
    \end{subfigure}
    \hfill
    \begin{subfigure}[c]{0.48\textwidth}
        \centering
        \includegraphics[width=\textwidth]{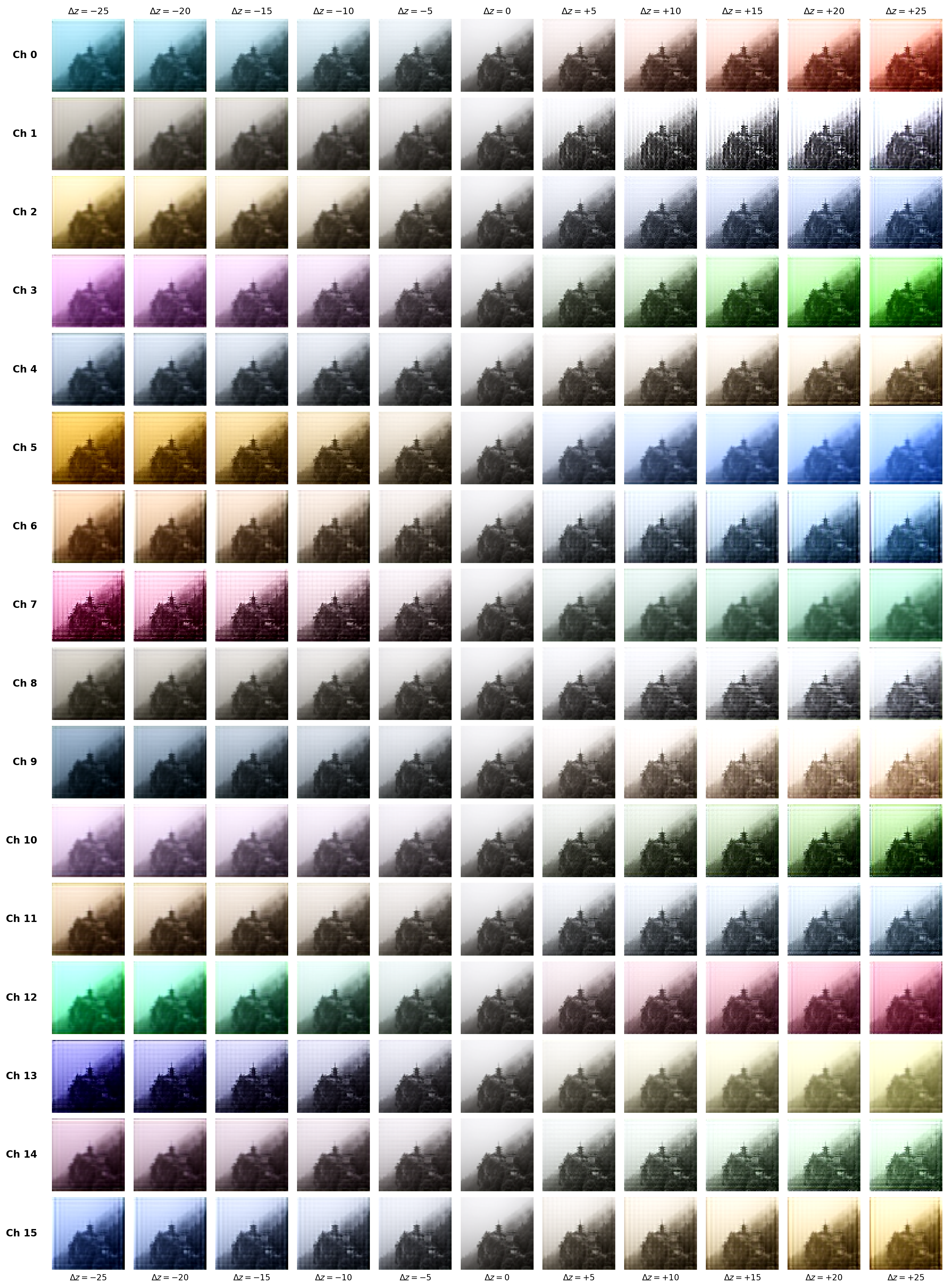}
        \caption{16-channel KL-VAE}
        \label{fig:shift_16ch}
    \end{subfigure}
    
    \caption{Visual impact of individual latent channel shifts ($\Delta z$) across different KL-VAE.}
    \label{fig:ae_channel_shifts}
\end{figure}

\begin{figure}[htbp]
    \centering
   
    \begin{subfigure}[c]{0.48\textwidth}
        \centering
        \includegraphics[width=\textwidth]{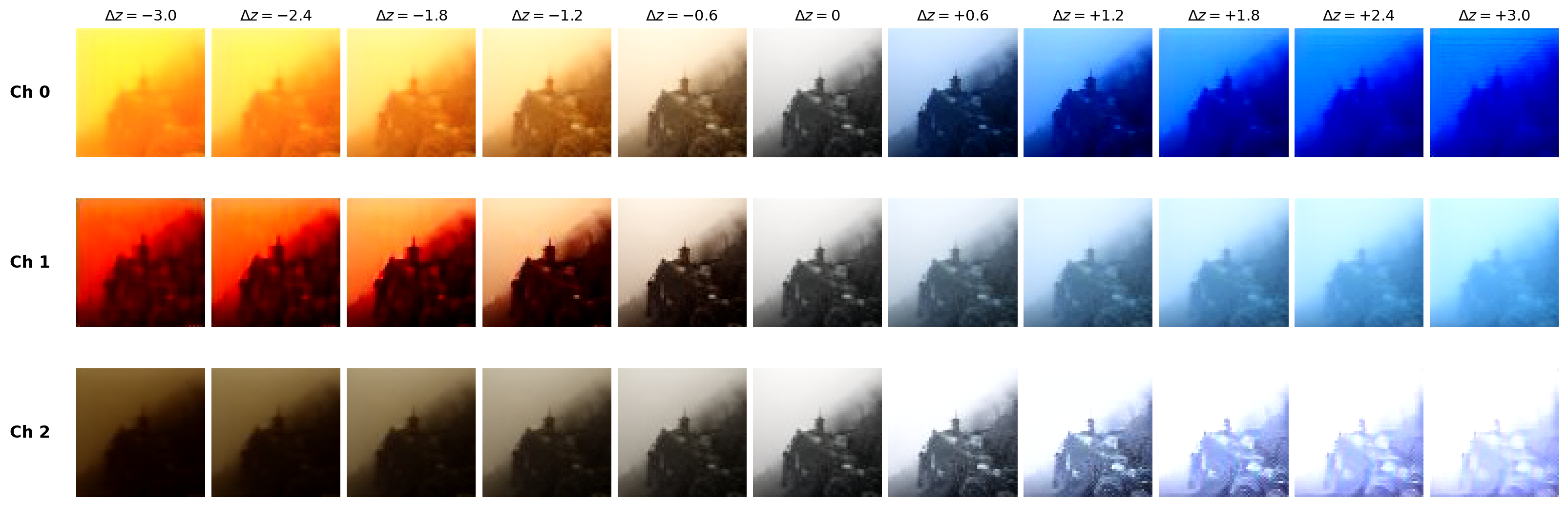}
        \caption{3-channel VQ-VAE}
        \label{fig:shift_3ch}
    \end{subfigure}
    \hfill
    \begin{subfigure}[c]{0.48\textwidth}
        \centering
        \includegraphics[width=\textwidth]{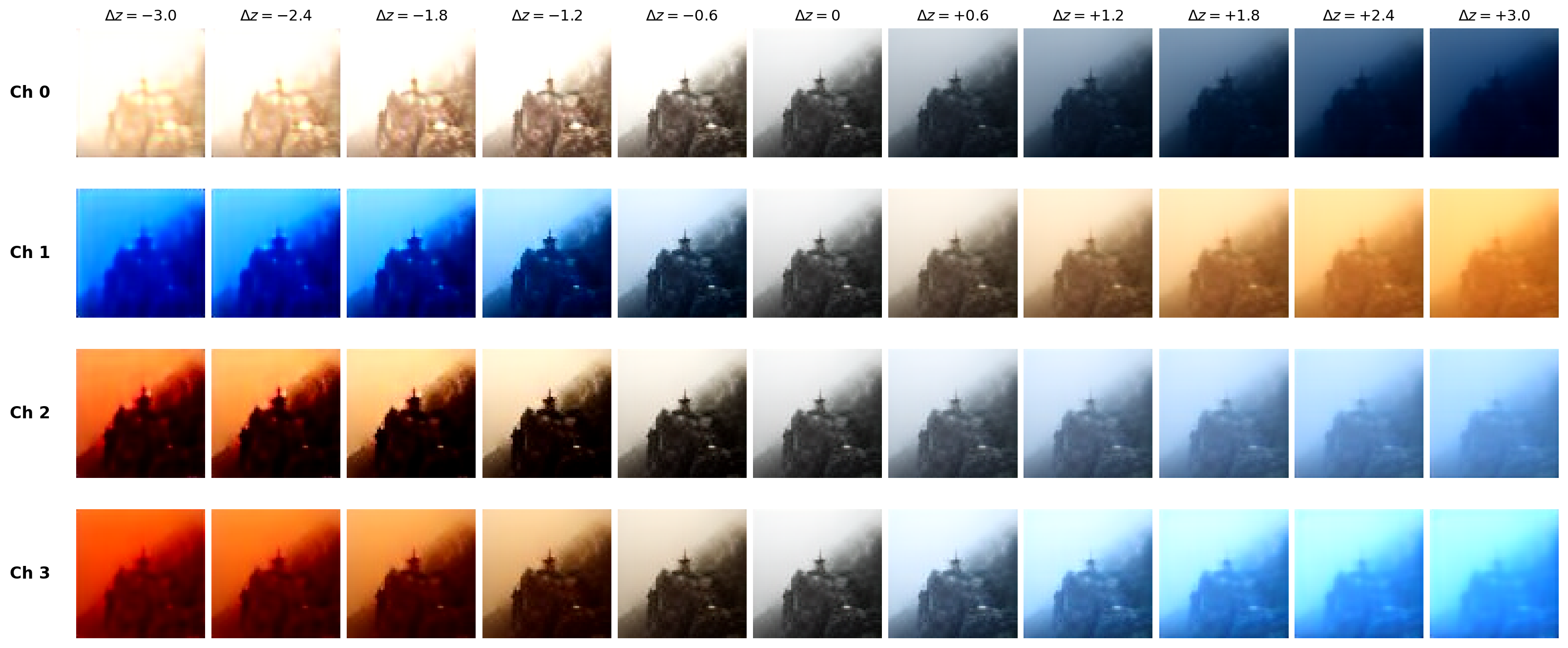}
        \caption{4-channel VQ-VAE}
        \label{fig:shift_4ch}
    \end{subfigure}
    
    \vspace{0.4cm}

    \begin{subfigure}[c]{0.48\textwidth}
        \centering
        \includegraphics[width=\textwidth]{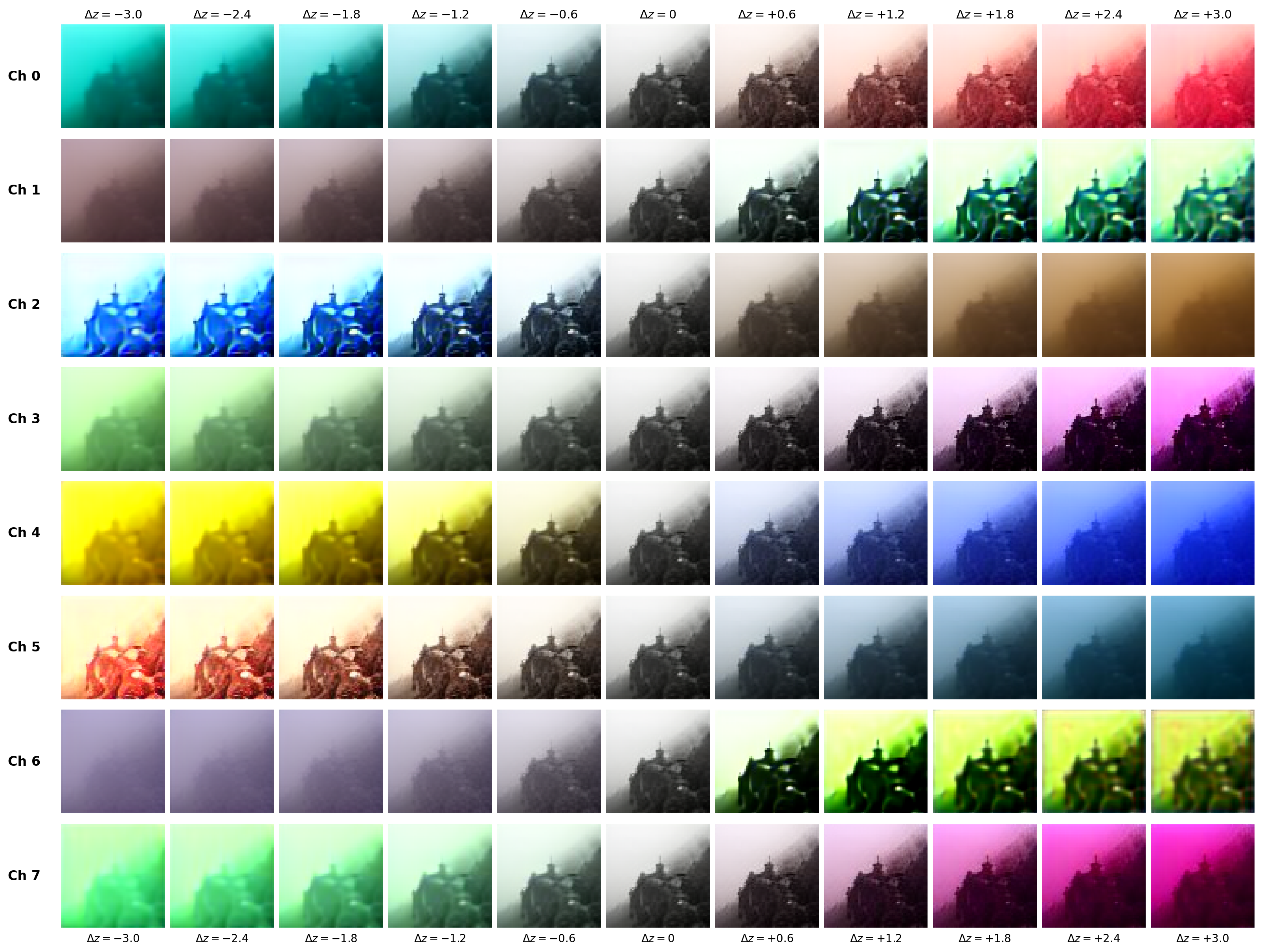}
        \caption{8-channel VQ-VAE}
        \label{fig:shift_8ch}
    \end{subfigure}
    \hfill
    \begin{subfigure}[c]{0.48\textwidth}
        \centering
        \includegraphics[width=\textwidth]{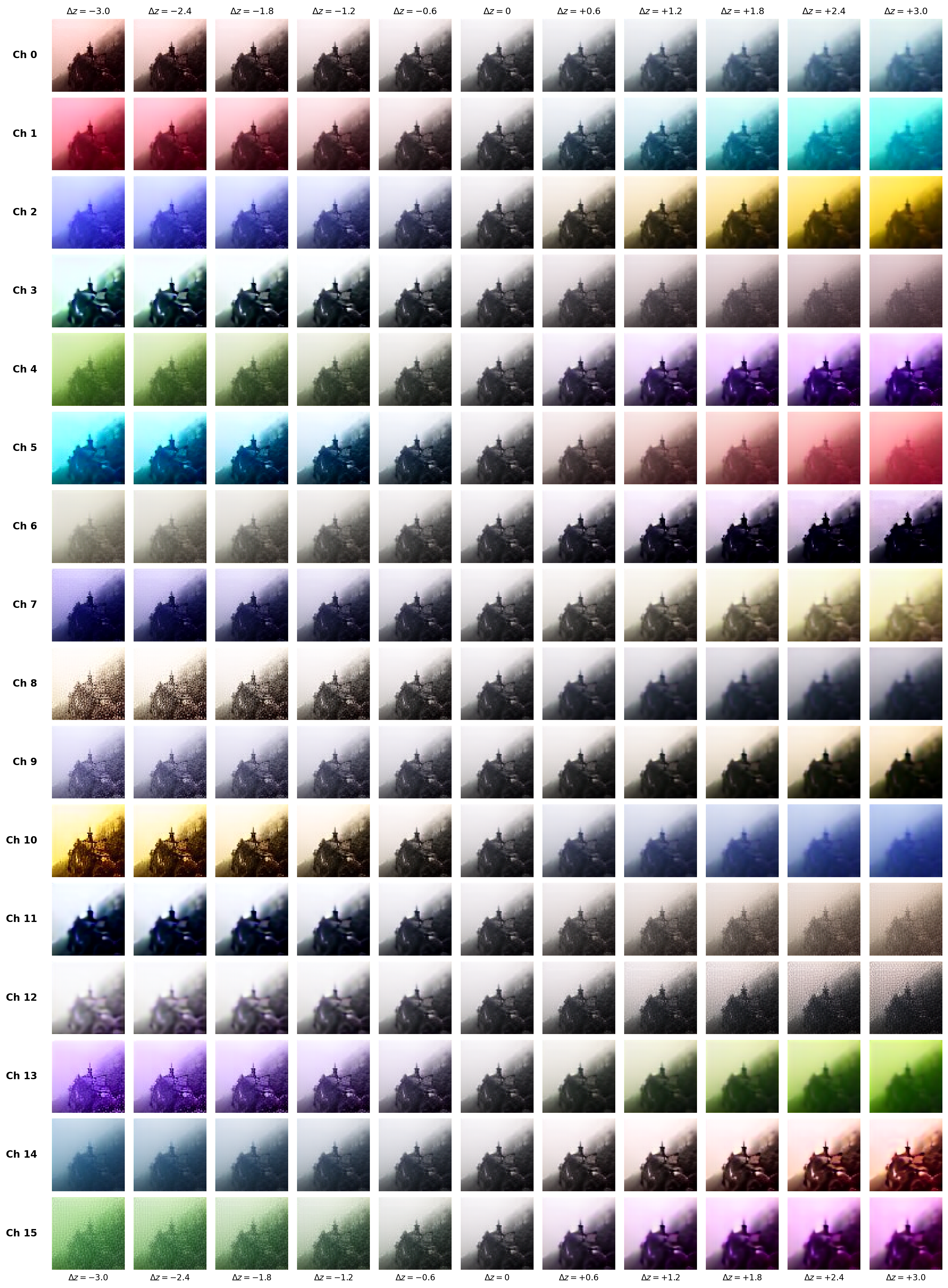}
        \caption{16-channel VQ-VAE}
        \label{fig:shift_16ch}
    \end{subfigure}
    
    \caption{Visual impact of individual latent channel shifts ($\Delta z$) across different VQ-VAE.}
    \label{fig:ae_channel_shifts}
\end{figure}

\subsubsection{Latent Diffusion Models}
\begin{figure}[htbp]
    \centering

    \begin{minipage}[c]{0.45\textwidth}
        \begin{subfigure}{\textwidth}
            \centering
            \includegraphics[width=\textwidth]{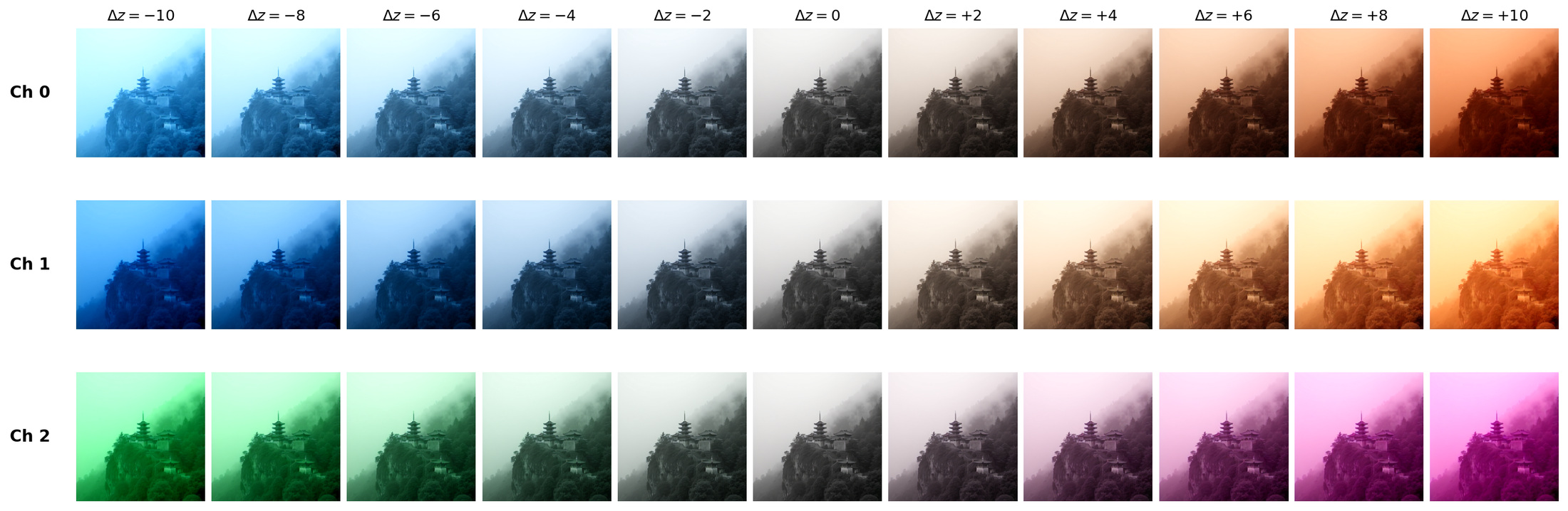}
            \caption{KL-f4 (3-channel)}
            \label{fig:kl_f4_3ch}
        \end{subfigure}
        
        \vspace{0.6cm}
        
        \begin{subfigure}{\textwidth}
            \centering
            \includegraphics[width=\textwidth]{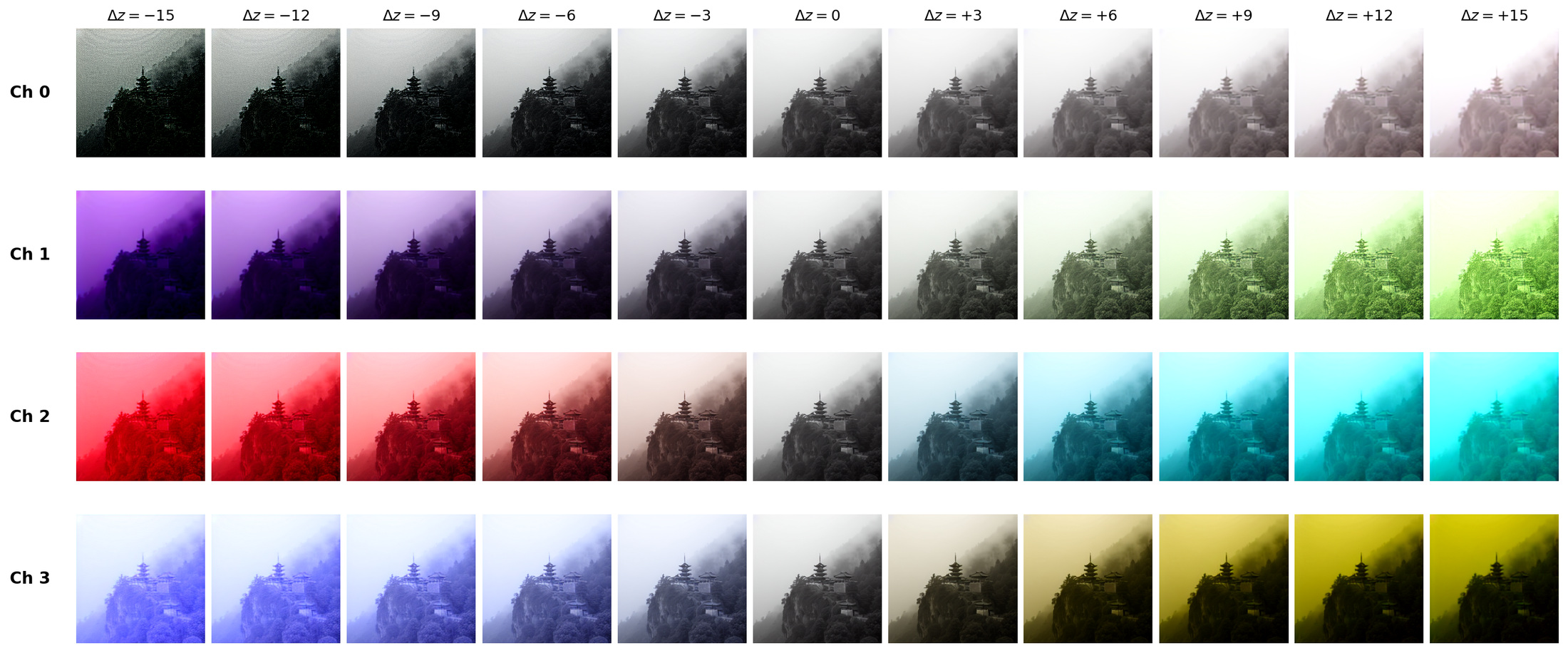}
            \caption{KL-f8 (4-channel)}
            \label{fig:kl_f8_4ch}
        \end{subfigure}
    \end{minipage}
    \hfill

    \begin{minipage}[c]{0.50\textwidth}
        \begin{subfigure}{\textwidth}
            \centering
            \includegraphics[width=\textwidth]{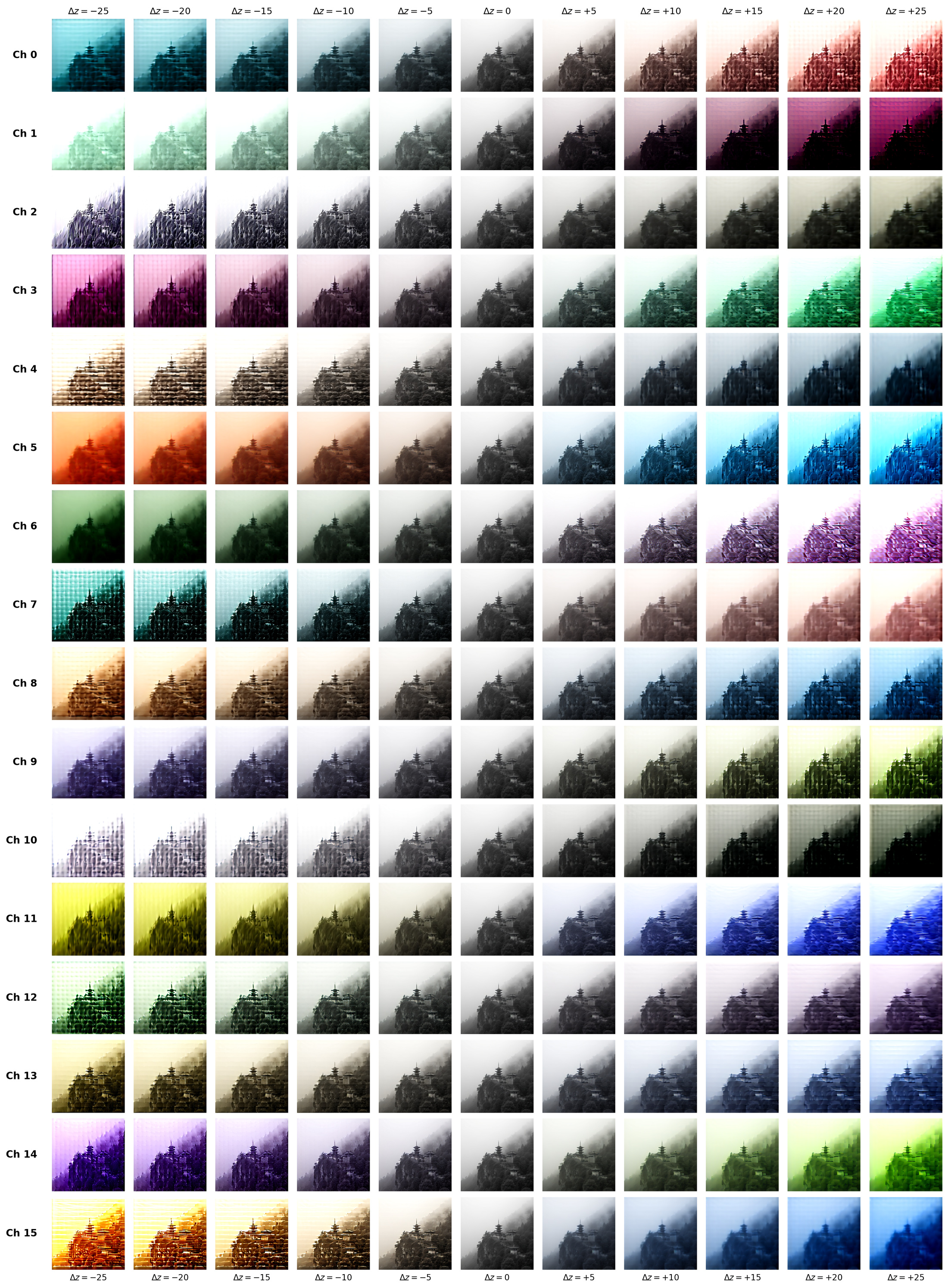}
            \caption{KL-f16 (16-channel)}
            \label{fig:kl_f16_16ch}
        \end{subfigure}
    \end{minipage}
    
    \caption{Latent channel shifts applied to KL-regularized VAEs with varying downsampling factors (f4, f8, f16).}
    \label{fig:kl_channel_shifts}
\end{figure}

\begin{figure}[htbp]
    \centering
   
    \begin{subfigure}[c]{0.48\textwidth}
        \centering
        \includegraphics[width=\textwidth]{figures/shift/local_train/ae/ae_3ch_channel_shifts.png}
        \caption{VQ-f4 (3-channel)}
        \label{fig:vq_f4_3ch}
    \end{subfigure}
    \hfill
    \begin{subfigure}[c]{0.48\textwidth}
        \centering
        \includegraphics[width=\textwidth]{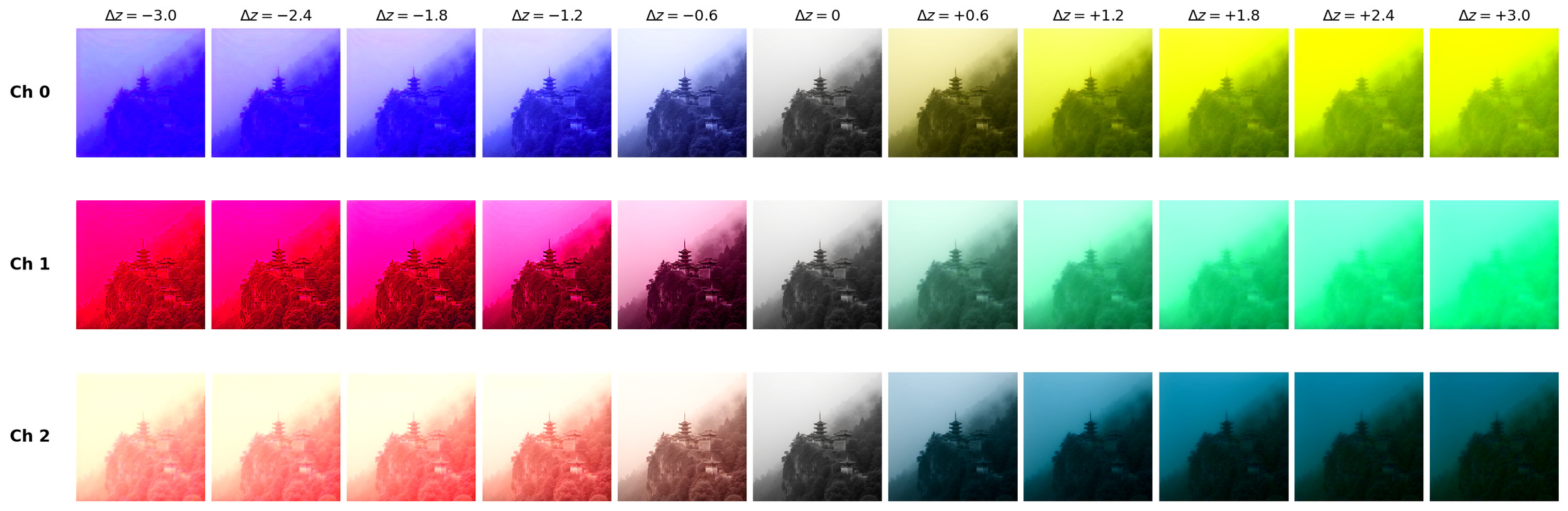}
        \caption{VQ-f4-noattn (3-channel)}
        \label{fig:vq_f4_noattn_3ch}
    \end{subfigure}
    
    \vspace{0.4cm}

    \begin{subfigure}[c]{0.48\textwidth}
        \centering
        \includegraphics[width=\textwidth]{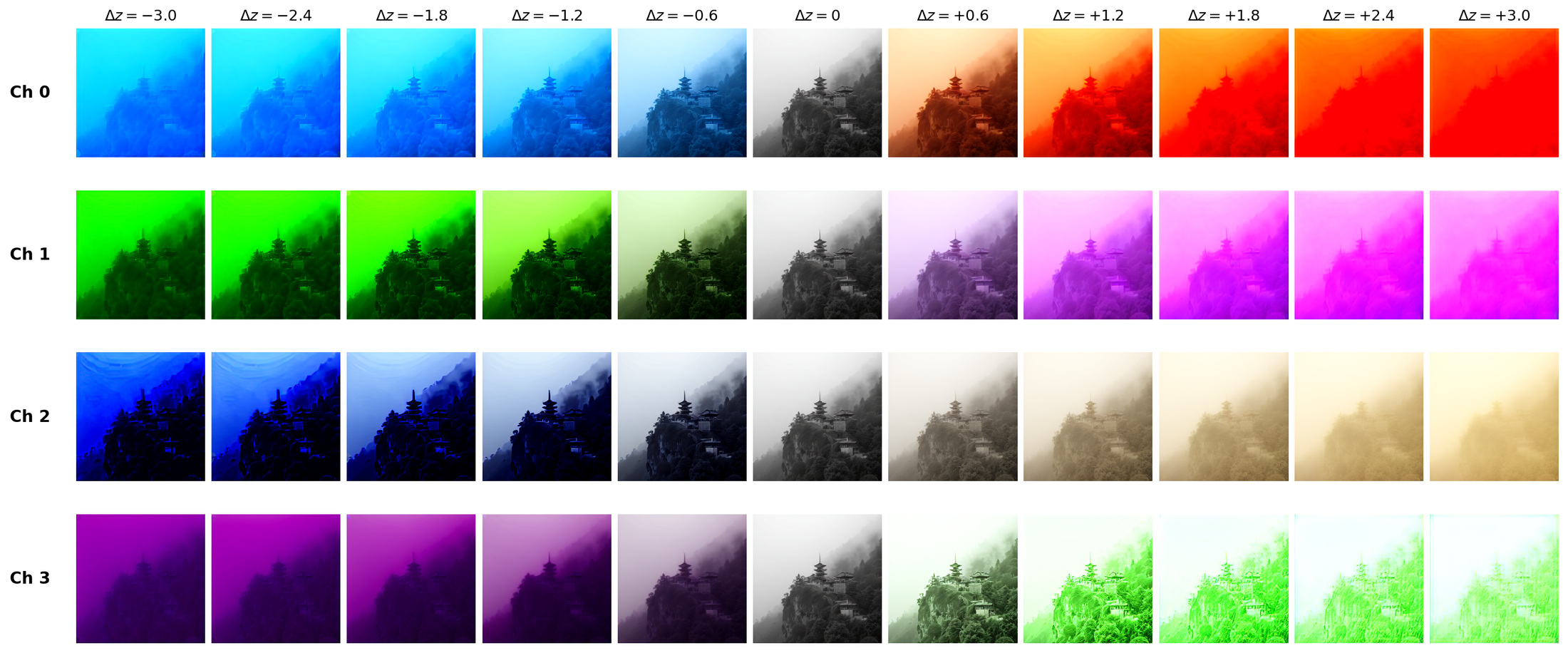}
        \caption{VQ-f8 (4-channel)}
        \label{fig:vq_f8_4ch}
    \end{subfigure}
    \hfill
    \begin{subfigure}[c]{0.48\textwidth}
        \centering
        \includegraphics[width=\textwidth]{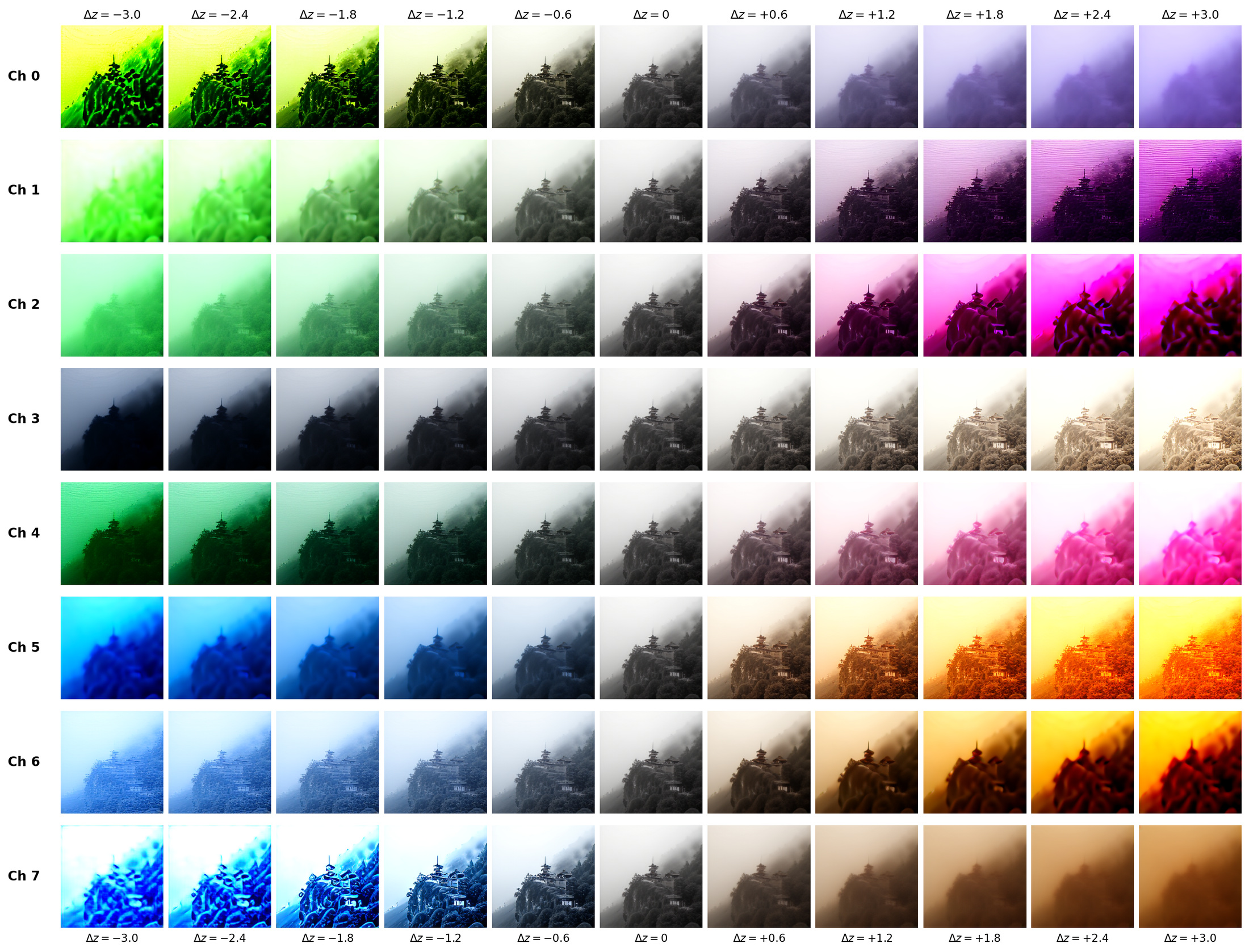}
        \caption{VQ-f16 (8-channel)}
        \label{fig:vq_f16_8ch}
    \end{subfigure}
    
    \caption{Latent channel shifts across Vector-Quantized (VQ) VAEs.}
    \label{fig:vq_channel_shifts}
\end{figure}

Building upon our local baselines, we extend our latent perturbation analysis to the pretrained autoencoders commonly utilized in LDMs~\citep{rombach2022high}. Specifically, we isolate and shift individual latent channels across both KL-VAE (Figure~\ref{fig:kl_channel_shifts}) and VQ-VAE (Figure~\ref{fig:vq_channel_shifts}) architectures.

\subsubsection{Foundation Models}
\begin{figure}[htbp]
    \centering

    \begin{subfigure}[c]{0.48\textwidth}
        \centering
        \includegraphics[width=\textwidth]{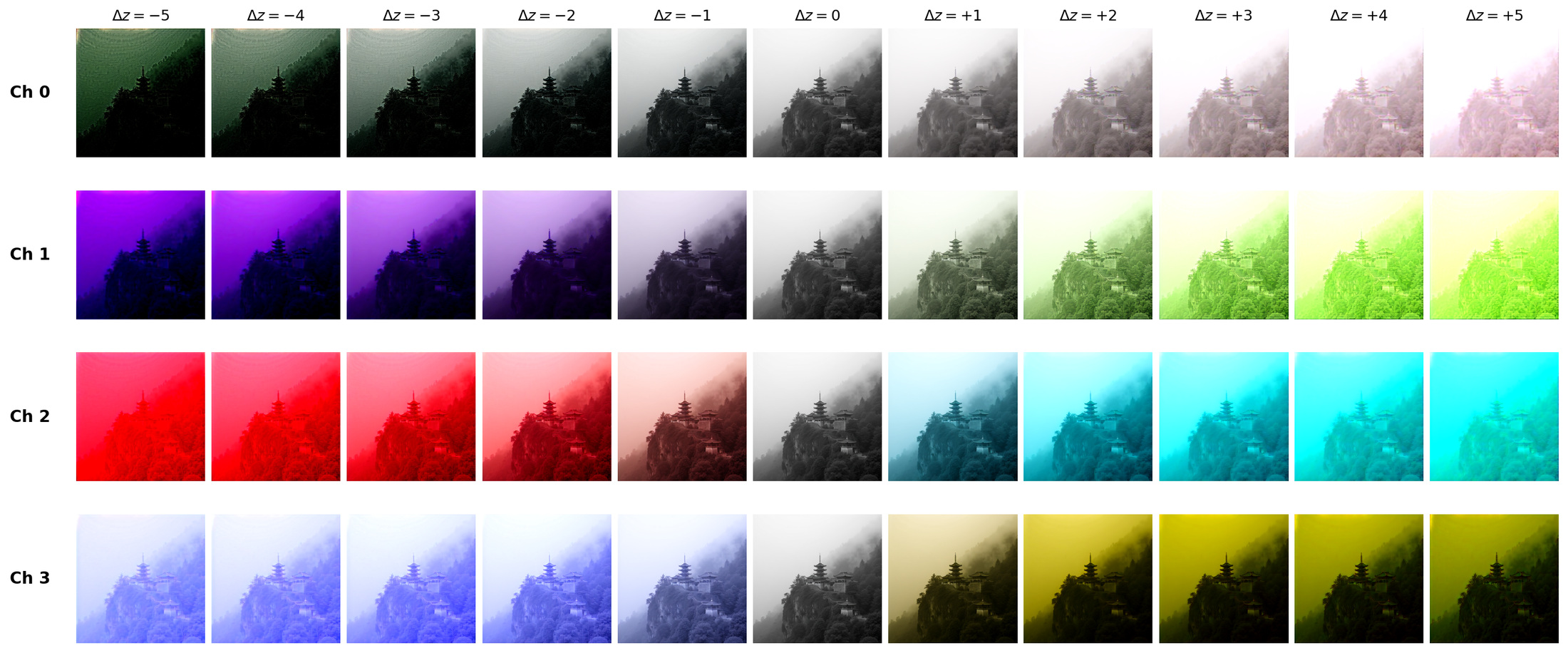}
        \caption{SD 1.5 (4-channel)}
        \label{fig:shift_sd15}
    \end{subfigure}
    \hfill
    \begin{subfigure}[c]{0.48\textwidth}
        \centering
        \includegraphics[width=\textwidth]{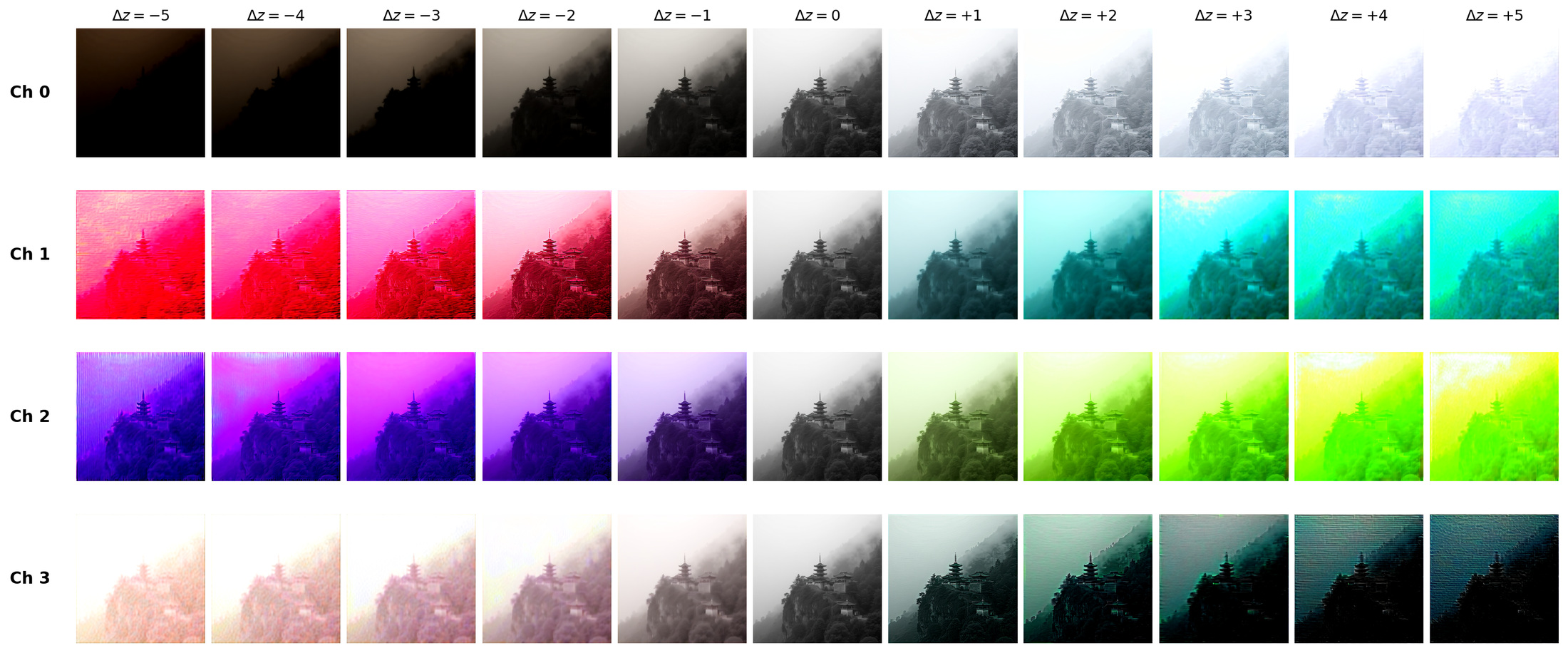}
        \caption{SDXL (4-channel)}
        \label{fig:shift_sdxl}
    \end{subfigure}
    
    \vspace{0.4cm}

    \begin{subfigure}[c]{0.48\textwidth}
        \centering
        \includegraphics[width=\textwidth]{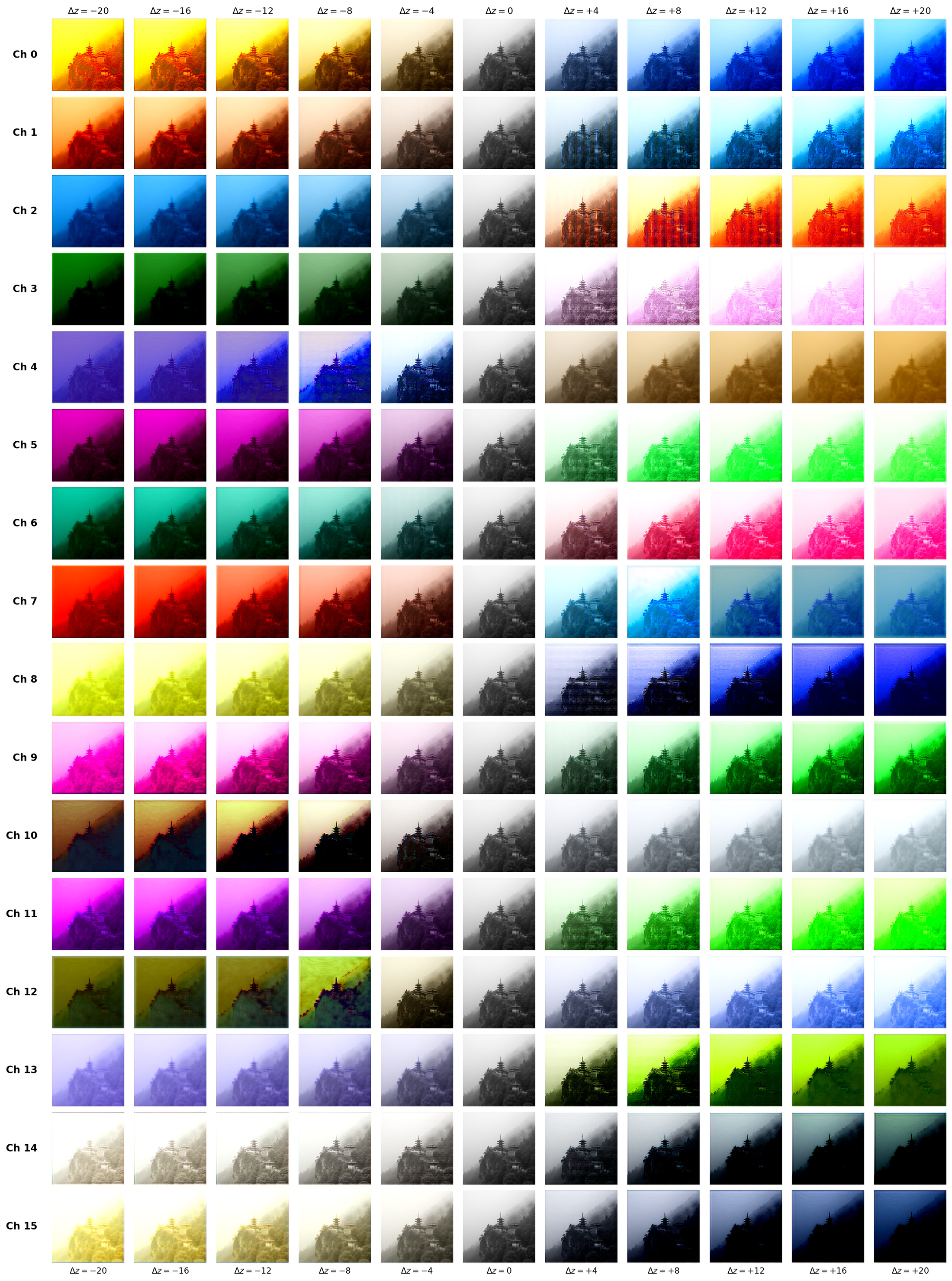}
        \caption{SD 3 (16-channel)}
        \label{fig:shift_sd3}
    \end{subfigure}
    \hfill
    \begin{subfigure}[c]{0.48\textwidth}
        \centering
        \includegraphics[width=\textwidth]{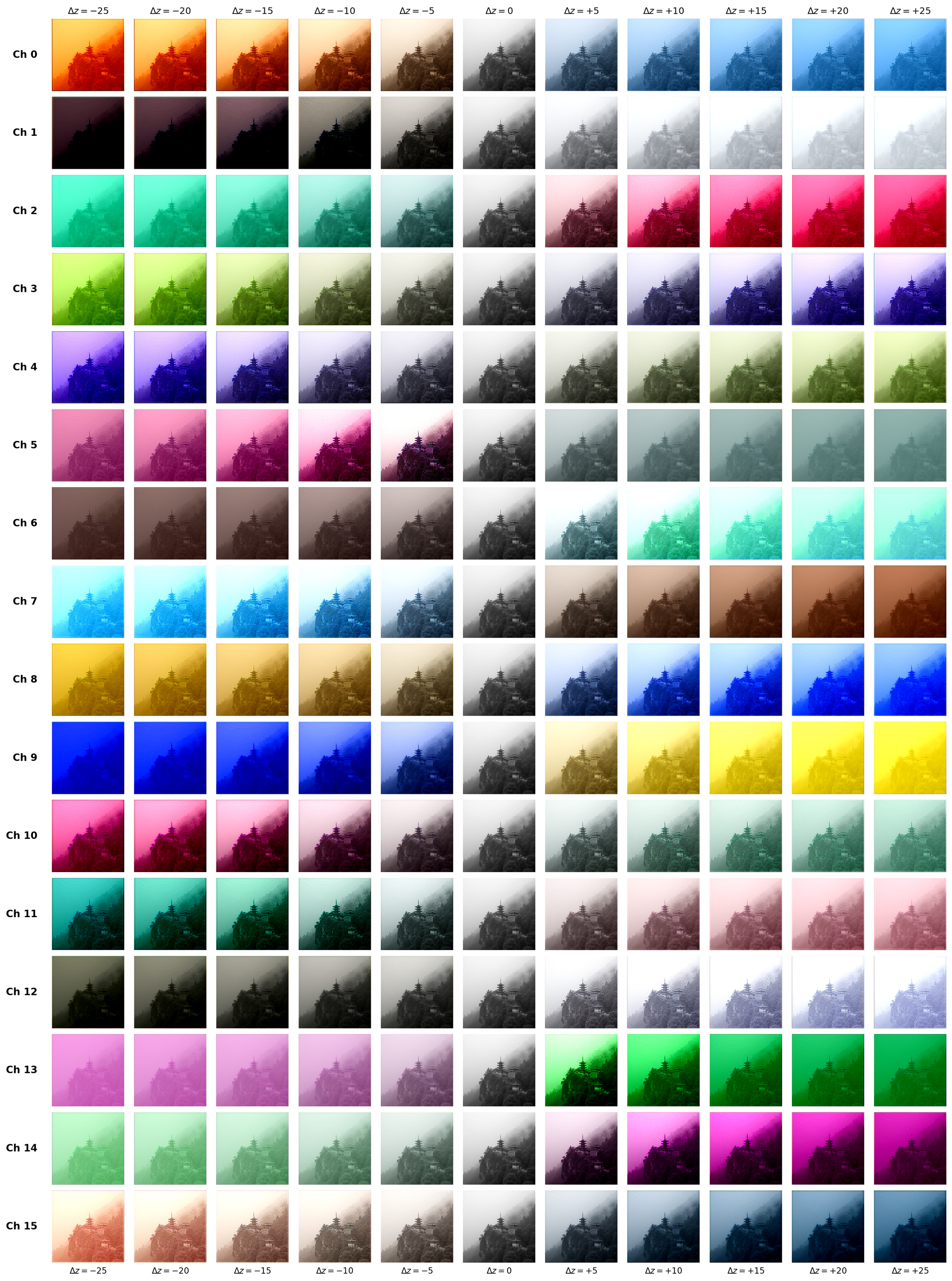}
        \caption{FLUX (16-channel)}
        \label{fig:shift_flux}
    \end{subfigure}
    
    \caption{Latent channel shifts in state-of-the-art foundation models. The top row compares the 4-channel autoencoders of SD1.5 and SDXL. The bottom row illustrates the expanded 16-channel latent spaces of SD3 and FLUX.1.}
    \label{fig:foundation_shifts}
\end{figure}

\begin{figure}[htbp]
    \centering
    \includegraphics[width=0.85\textwidth, height=0.9\textheight, keepaspectratio]{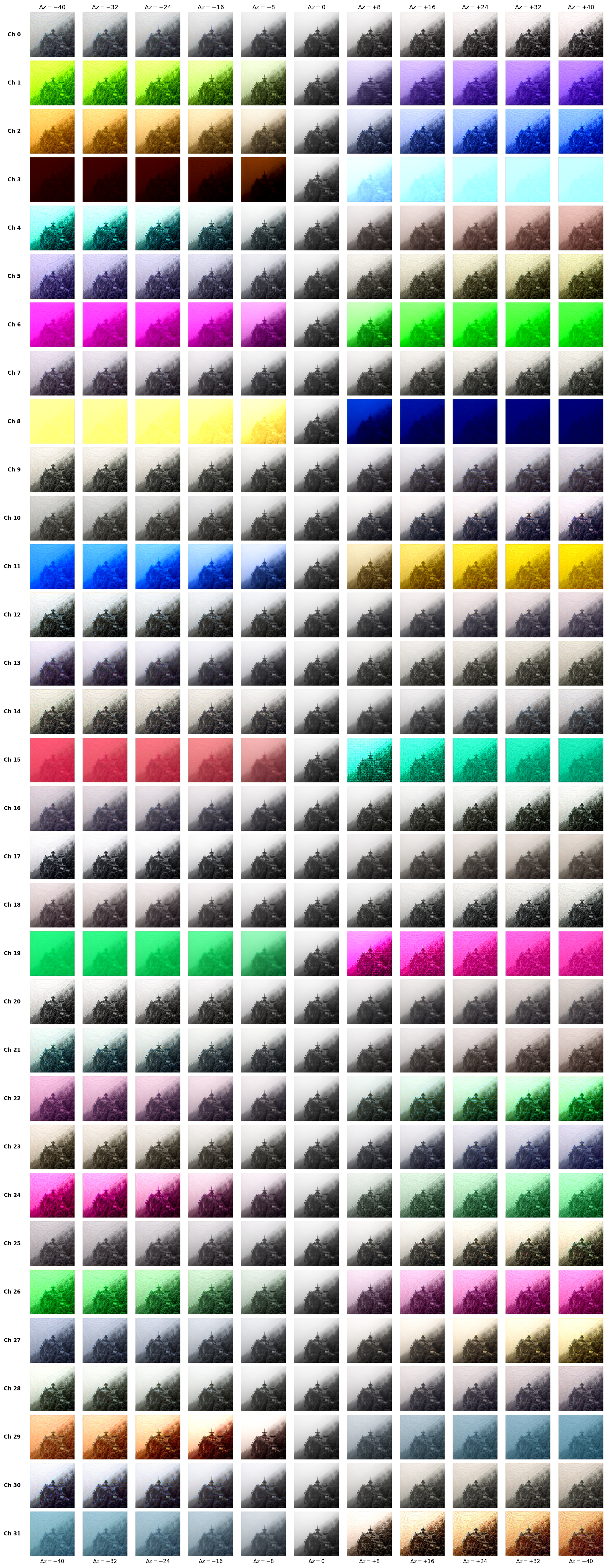}
    \caption{Extended latent channel shifts for the 32-channel FLUX2 autoencoder.}
    \label{fig:flux2_shifts}
\end{figure}

Finally, we investigate the latent space architectures of foundation models. We apply our latent perturbation analysis to the autoencoders of SD1.5, SDXL, SD3, FLUX, and FLUX2, as illustrated in Figure~\ref{fig:foundation_shifts} and Figure~\ref{fig:flux2_shifts}.

\subsection{Similar color subspaces emerge in different datasets.}\label{larger}

\begin{figure*}[t]
    \centering
        \centering
   
        \includegraphics[width=0.9\textwidth]{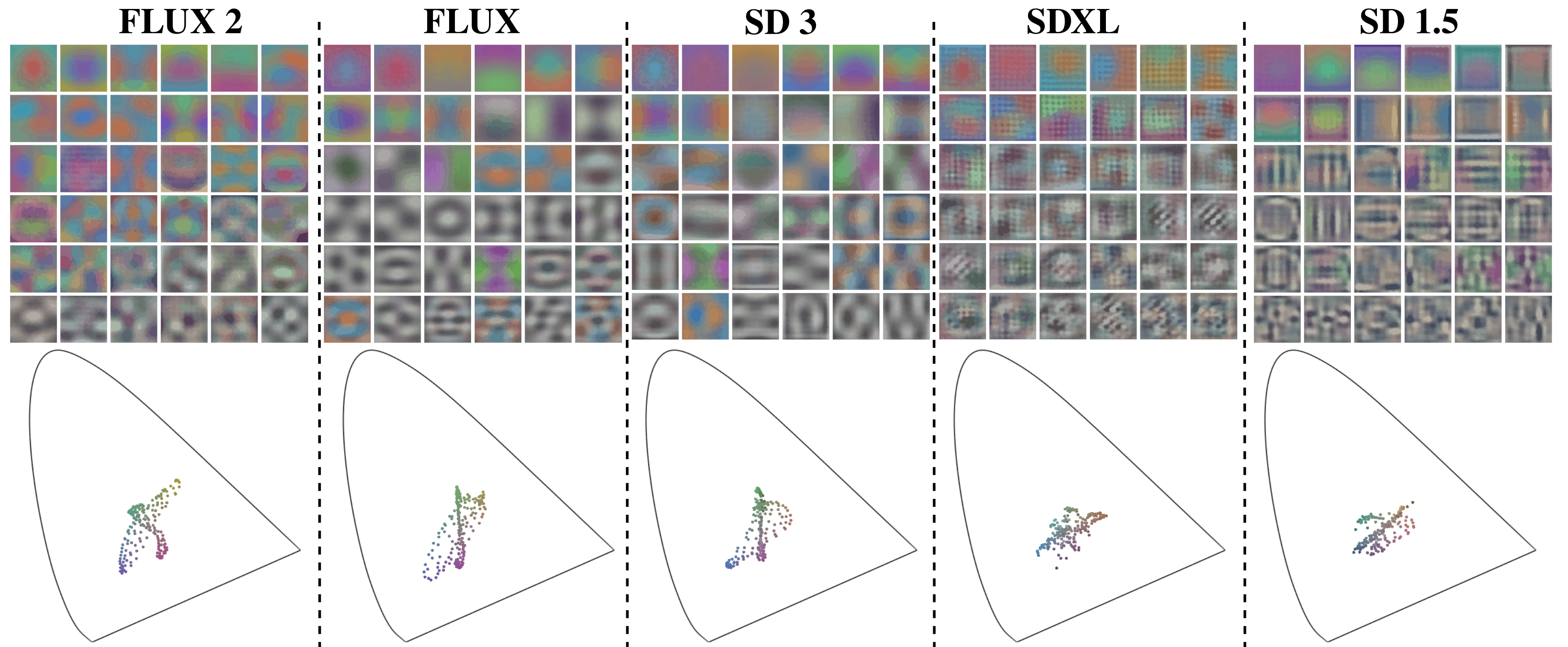}
         \caption{The first singular vectors of the linear approximation of VAEs and their corresponding CIE chromaticity directions across generative models used with $64\times64$ images: Flux.dev~2, Flux.dev~1, (c) SD 1.5, (d) SDXL, and SD 3. Each subplot visualizes the singular vectors ordered by their chromatic variance alongside their color content in the CIE chromatic diagram.}\label{fig:vae_eigenfunctions_grid_large}  
\end{figure*}

 Fig.~\ref{fig:vae_eigenfunctions_grid_large}  shows the singular vectors and color content of the latent space of $M_{enc}$ obtained from a set of images with substantially different resolutions: $10^5$ images of size $64\times64$. In this case, for a cleaner visualization, a standard variance minimization algorithm~\citep{Wu91} has been applied to select the colors shown. Qualitative similarity of these results with those of Fig.~\ref{fig:vae_eigenfunctions_grid} suggests that color opponent subspaces emerge consistently for different sets of natural images and resolutions.

\subsection{Faithfulness of the linear approximations.}\label{faith}

In this Appendix, we quantify the validity of the linear approximation of VAE's encoder and decoder according to the methods proposed by \citet{gomez2020color} for regular AEs.
In particular, here we compute three metrics for each VAE:
\emph{Fract.\ Lin.\ Resp.} is the fraction of the energy of the network response
explained by the linear approximation. 
It is computed from the RMSE between the linear and the actual (nonlinear) responses, given in relative units with regard to the square root of the energy of the actual response. 

\emph{Error NonLin.} and \emph{Error Linear} are the corresponding reconstruction errors of the VAE and of its linear approximation,
each expressed as the fraction of the energy of the original image.

Table~\ref{table:lin_resp_large} shows such metrics and the average across all VAEs.
Results show that on average, the linear approximations explain a large fraction (about $65 \%$) of the nonlinear behavior. 
Moreover, the magnitudes of the reconstruction error by the VAEs and by their linear approximations are similar (about $12\%$ of the energy of the original signal).

These results imply that the singular vectors shown in this work are a meaningful description of what the VAEs are sensitive to, both in terms of texture and color directions.

\begin{table}[t]
\centering
\begin{tabular}{lcccccc}
\multicolumn{7}{c}{$64\times64$ images} \\
\hline
 & FLUX 2 & FLUX & SD3 & SDXL & SD1.5  & \textbf{AVERAGE}\\
\hline
\multicolumn{7}{c}{\textbf{\textit{Lin. Enc. + Lin. Dec}}} \\
\hline
Fract. Lin. Resp.   & 0.67 & 0.73 & 0.72 & 0.71 & 0.74 & \textbf{0.71 $\pm$ 0.02}\\
Error No Linear     & 0.02 & 0.03 & 0.05 & 0.08 & 0.08 & \textbf{0.05 $\pm$ 0.02}\\
Error Linear        & 0.11 & 0.09 & 0.09 & 0.09 & 0.09 & \textbf{0.09 $\pm$ 0.01}\\
\hline
\multicolumn{7}{c}{\textbf{\textit{Lin. Enc. + NonLin. Dec}} } \\
\hline
Fract. Lin. Resp.   & 0.67 & 0.74 & 0.58 & 0.61 & 0.69 & \textbf{0.66 $\pm$ 0.06}\\
Error No Linear     & 0.02 & 0.03 & 0.05 & 0.08 & 0.08 &\textbf{0.05 $\pm$ 0.02}\\
Error Linear        & 0.11 & 0.08 & 0.13 & 0.13 & 0.10 & \textbf{0.11 $\pm$ 0.02}\\
\hline
\multicolumn{7}{c}{\textbf{\textit{NonLin. Enc. + Lin. Dec}}} \\
\hline
Fract. Lin. Resp.   & 0.70 & 0.76 & 0.75 & 0.71 & 0.72 & \textbf{0.73 $\pm$ 0.02}\\
Error No Linear     & 0.02 & 0.03 & 0.05 & 0.08 & 0.08 & \textbf{0.05 $\pm$ 0.02} \\
Error Linear        & 0.10 & 0.08 & 0.08 & 0.09 & 0.09 & \textbf{0.09 $\pm$ 0.01}\\
\hline\\[0.2cm]
\multicolumn{7}{c}{$16\times16$ images} \\
\hline
 & FLUX 2 & FLUX & SD3 & SDXL & SD1.5   & \textbf{AVERAGE}\\
\hline
\multicolumn{7}{c}{\textbf{\textit{Lin. Enc. + Lin. Dec}}} \\
\hline
Fract. Lin. Resp.   & 0.83 & 0.80 & 0.69 & 0.55 & 0.73 & \textbf{0.72 $\pm$ 0.10} \\
Error No Linear     & 0.05 & 0.09 & 0.37 & 0.27 & 0.19 & \textbf{0.19 $\pm$ 0.10} \\
Error Linear        & 0.05 & 0.06 & 0.09 & 0.14 & 0.08 & \textbf{0.08 $\pm$ 0.03} \\
\hline
\multicolumn{7}{c}{\textbf{\textit{Lin. Enc. + NonLin. Dec}} } \\
\hline
Fract. Lin. Resp.   & 0.67 & 0.68 & -0.03 & 0.23 & 0.45 & \textbf{0.40 $\pm$ 0.30} \\
Error No Linear     & 0.05 & 0.09 & 0.37 & 0.27 & 0.19 & \textbf{0.19 $\pm$ 0.10} \\
Error Linear        & 0.09 & 0.10 & 0.31 & 0.23 & 0.17 & \textbf{0.18 $\pm$ 0.09} \\
\hline
\multicolumn{7}{c}{\textbf{\textit{NonLin. Enc. + Lin. Dec}}} \\
\hline
Fract. Lin. Resp.   & 0.78 & 0.75 & 0.47 & 0.41 & 0.60 & \textbf{0.60 $\pm$ 0.20} \\
Error No Linear     & 0.05 & 0.09 & 0.37 & 0.27 & 0.19 & \textbf{0.19 $\pm$ 0.10} \\
Error Linear        & 0.07 & 0.08 & 0.20 & 0.21 & 0.12 & \textbf{0.14 $\pm$ 0.04} \\
\hline
\end{tabular}
\caption{Linear approximation metrics across all models, configurations and datasets.}
\label{table:lin_resp_large}
\end{table}

\subsection{T2I models fail at numerical color precision}

\begin{figure}[h]
\begin{center}
\includegraphics[width=1\linewidth]{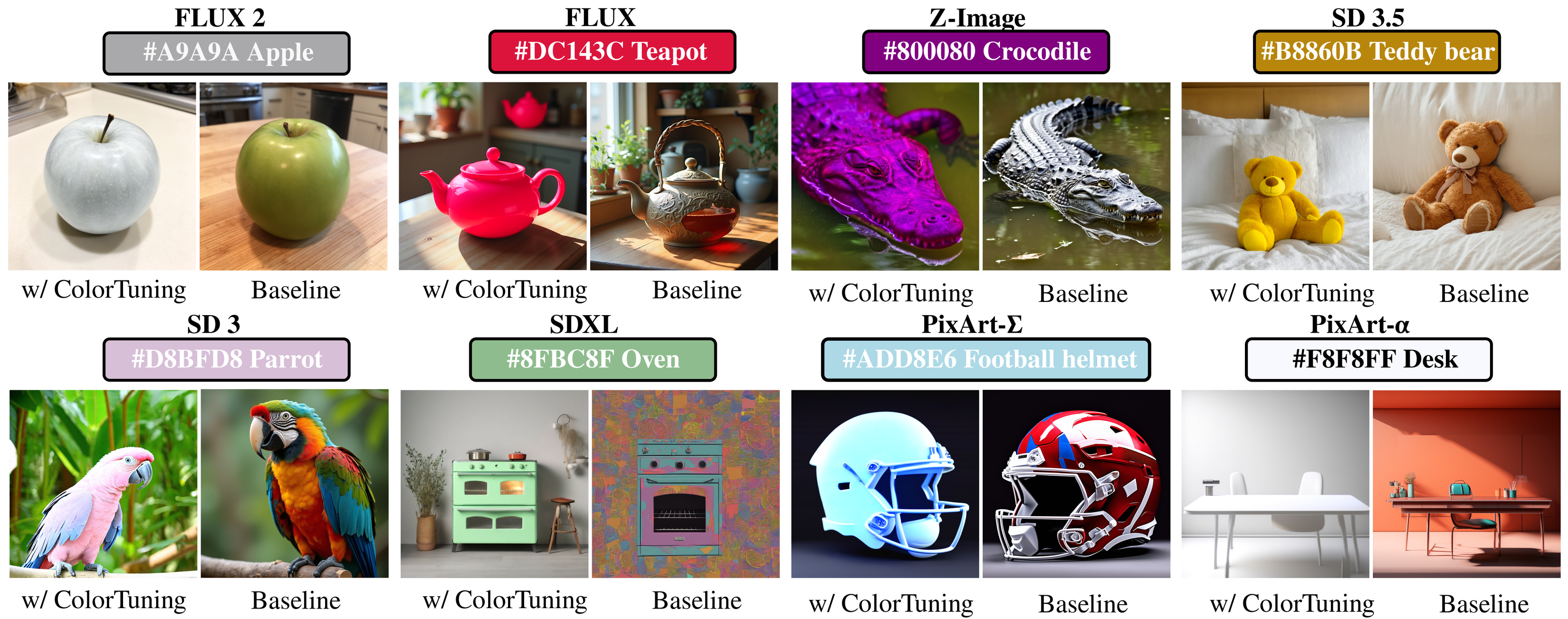}
\end{center}
\caption{T2I models fail at numerical color precision.}
\label{fig:errors_gen}
\end{figure}

\subsection{Pipelines}
\label{app:pipelines}

We evaluate our method across a diverse set of state-of-the-art text-to-image models operating on distinct VAE latent spaces. \method is implemented on models with 4-channel (SDXL-Base-1.0), 16-channel (SD3-Medium, SD3.5-Medium, Z-Image, FLUX.1-dev), and FLUX.2-dev with 32-channel. To compute the spatial conditioning mask, we use SAM-3 (\texttt{facebook/sam3}) with the target object from the prompt.

\begin{figure}[t]
\begin{center}
\includegraphics[width=1\linewidth]{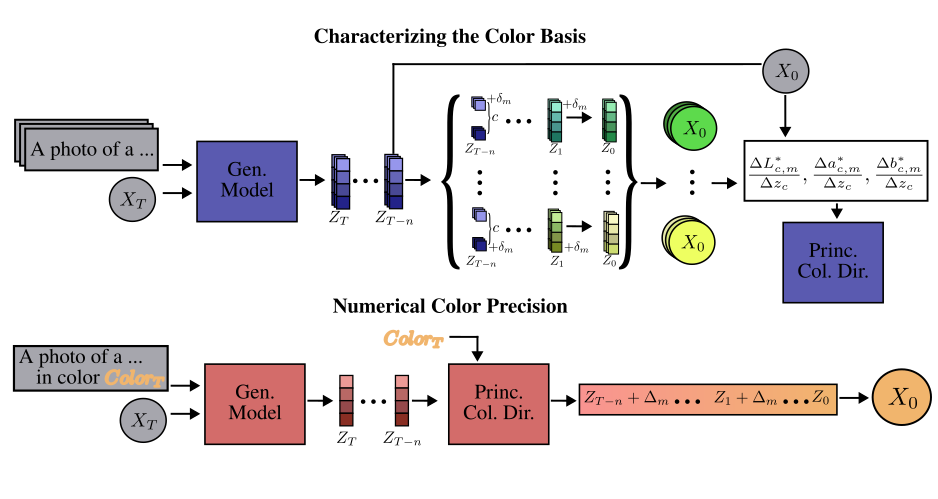}
\end{center}
\caption{Overview of our method. \textbf{Top:} characterizing the color basis of a VAE (Sec.~\ref{sec:color_basis}). Each latent channel is perturbed inside the region of a set of reference objects, and the resulting shifts in CIELAB form a sensitivity matrix, whose SVD gives the luminance and opponent-color directions. This is computed once per VAE. \textbf{Bottom:} \method\ at inference (Sec.~\ref{sec:numerical}). The numerical code in the prompt is replaced by the nearest color name. At the gate step, the predicted clean latent $\hat{z}_0$ is decoded, the object is segmented with SAM~3, and its mean color is measured. The MLP maps this color and the target to a displacement along the color basis, which is added inside the object mask over the remaining steps.}
\label{fig:method}
\end{figure}

\subsubsection{Temporal Injection Schedules}
Perturbations along the three principal PCA color axes are applied over an active generation window $[\tau_{\text{gate}}, 1.0]$. These perturbations are modulated by the spatial mask and a temporal schedule $w(t)$. The specific gate step $\tau_{\text{gate}}$ and the optimal schedule profile for each model are detailed in Table \ref{tab:pipeline_hyperparams}. We employ three distinct temporal profiles:
\begin{itemize}[leftmargin=*]
\item \textbf{Ascending Ramp} ($w(t) = \frac{t - \tau_{\text{gate}}}{1 - \tau_{\text{gate}}}$): Gradually increases perturbation intensity.
\item \textbf{Triangular Profile}: Reaches peak intensity at the mid-window to preserve fine structural textures.
\item \textbf{Constant Profile} ($w(t) = 1.0$): Applies uniform injection across the active window.
\end{itemize}

\subsubsection{Shift Predictor Network Architecture}
To map the desired color shifts to PCA magnitude vectors $\mathbf{m} = (m_1, m_2, m_3)^\top$, we train a Multi-Layer Perceptron with residual connections.
\begin{itemize}[leftmargin=*]
\item \textbf{Input Features:} The network ingests a 15-dimensional perceptual feature vector combining initial and target CIELAB coordinates, chroma values, and periodic sine/cosine hue representations.
\item \textbf{Architecture:} An input projection layer (15 to 256 dimensions) is followed by two residual blocks (each containing two 256-dimensional linear layers, LayerNorm, SiLU activation, and a dropout of $p = 0.05$). A final bottleneck projects the features down to 3 dimensions to predict the shift.
\item \textbf{Training Setup:} The network is optimized using AdamW (learning rate $\eta = 10^{-3}$, weight decay $10^{-4}$) with a Cosine Annealing schedule (minimum learning rate $\eta_{\min} = 10^{-5}$). We train for 80 epochs using a Smooth L1 loss ($\beta = 0.05$), a batch size of 64, and an early stopping patience of 15 epochs.
\end{itemize}

\begin{table}[htpb]
\centering
\footnotesize
\caption{Pipeline Specifications and Hyperparameters .}
\label{tab:pipeline_hyperparams}
\setlength{\tabcolsep}{3.8pt}
\begin{tabular}{lccccc}
\toprule
\textbf{Model}  & \textbf{Steps} & \textbf{CFG}  & \textbf{Gate $\tau_{\text{gate}}$} & \textbf{Temporal Schedule} \\
\midrule
FLUX.1-dev   &    28 & 3.5 & 0.50 & Ascending  \\
FLUX.2-dev   &    28 & 3.5 & 0.65 & Ascending  \\
SD3-Medium   &    28 & 7.0 & 0.75 & Triangular \\
SD3.5-Medium &    28 & 4.5 & 0.60 & Ascending \\
SDXL-Base    &    30 & 5.0 & 0.40 & Constant  \\

Z-Image         & 28 & 4.0 & 0.60 & Constant \\
\bottomrule
\end{tabular}
\end{table}

\begin{algorithm}[h]
\caption{\method: training the displacement predictor}
\label{alg:training}
\begin{algorithmic}[1]
\Require frozen VAE $(E,D)$, generator, color basis $U \in \mathbb{R}^{C\times 3}$,
         prompt set $\mathcal{P}$, displacement set $\mathcal{M}$
\State $\mathcal{D} \gets \emptyset$
\For{$p \in \mathcal{P}$}
  \State $I_0 \gets \textsc{Generate}(p)$ \Comment{no intervention}
  \State $c_0 \gets \textsc{MeanLab}(I_0, \textsc{Sam3}(I_0, p))$
  \For{$m \in \mathcal{M}$}
    \State $I_m \gets \textsc{Generate}(p, \Delta z = U m)$
    \State $c_m \gets \textsc{MeanLab}(I_m, \textsc{Sam3}(I_m, p))$
    \State $\mathcal{D} \gets \mathcal{D} \cup \{((c_0, c_m), m)\}$
  \EndFor
\EndFor
\State train MLP $f_\psi: (c_i, c_t) \mapsto m$ on $\mathcal{D}$
\Ensure $f_\psi$
\end{algorithmic}
\end{algorithm}

\begin{algorithm}[h]
\caption{\method: inference}
\label{alg:inference}
\begin{algorithmic}[1]
\Require prompt $p$ with target color $c_t$, gate step $s$, total steps $T$,
         basis $U$, predictor $f_\psi$, proxy table $\mathcal{L}$
\State $p' \gets \textsc{ReplaceCode}(p, \textsc{NearestName}(c_t, \mathcal{L}))$
\State $z_T \sim \mathcal{N}(\mathbf{0}, I)$
\For{$t = T, \dots, 1$}
  \If{$t = s$}
    \State $\hat{z}_0 \gets$ Eq.~\ref{eq:x0_pred};\quad $I \gets D(\hat{z}_0)$
    \State $M \gets \textsc{Downsample}(\textsc{Sam3}(I, p'))$
    \State $c_i \gets \textsc{MeanLab}(I, M)$;\quad
           $\Delta z \gets U f_\psi(c_i, c_t)$
  \EndIf
  \State $z_{t-1} \gets \textsc{Step}(z_t, t)$
  \If{$t \leq s$}
    \State $z_{t-1} \gets z_{t-1} + w(t)\,(M \odot \Delta z)$
    \Comment{$w$ decays linearly to $0$}
  \EndIf
\EndFor
\Ensure $D(z_0)$
\end{algorithmic}
\end{algorithm}

\subsection{Saturation Control and Color Transfer}\label{app:further}
\begin{figure}[htpb]
    \centering
    \includegraphics[width=1\linewidth]{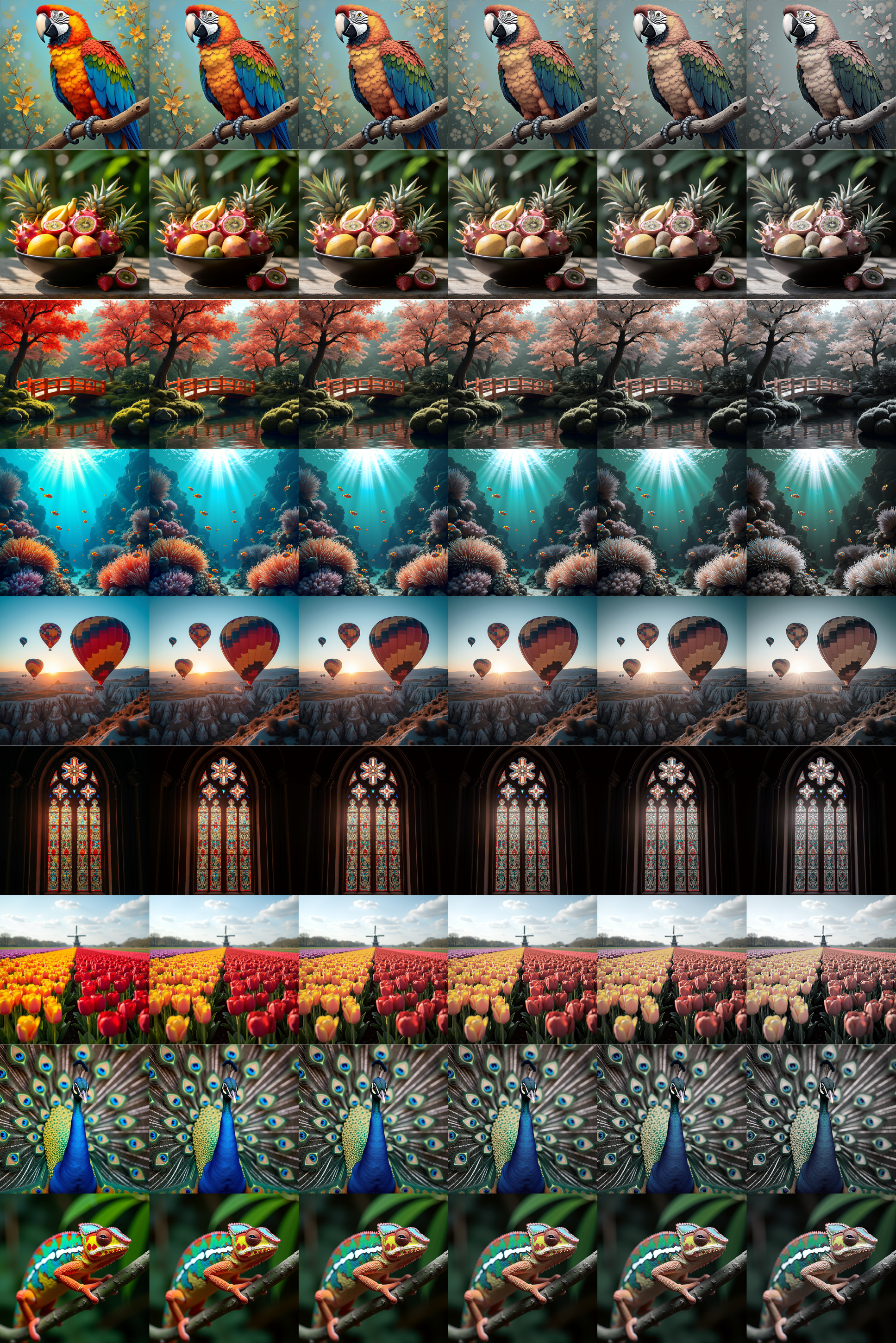}
    \caption{Qualitative results for saturation control.}
    \label{fig:sat_control}
\end{figure}

\textbf{Saturation Control.}
Saturation control continuously modulates the chroma of the generated scene or specific objects without altering their semantic structure or hue. In this setting, the text prompt requires no proxy injection, allowing the diffusion trajectory to proceed unconstrained until the gate step. At step $s$, we decode $\hat{z}_0$ and obtain the CIELAB color $C_i(x, y)= = (L_i(x, y), a_i(x, y), b_i(x, y))$ at each spatial location $(x, y)$ within the target region (or across the full image). For a desired chroma reduction factor $\alpha \in [0, 1]$, we preserve the lightness $L_i$ and hue angle while scaling chroma, setting the target color to $C_t = (L_i, (1 - \alpha) a_i, (1 - \alpha) b_i)$. The pretrained MLP takes $(C_i, C_t)$ and evaluates $(C_i(x, y),C_t(x, y))$ across the grid to predict a dense displacement map $m_{1}(x,y), m_{2}(x,y), m_{3}(x,y)$. Adding this displacement over the remaining steps pulls the generation into the restricted gamut within the latent space (see Figure~\ref{fig:sat_control}).
  
\begin{figure}[htpb]
    \centering
    \includegraphics[width=0.75\linewidth]{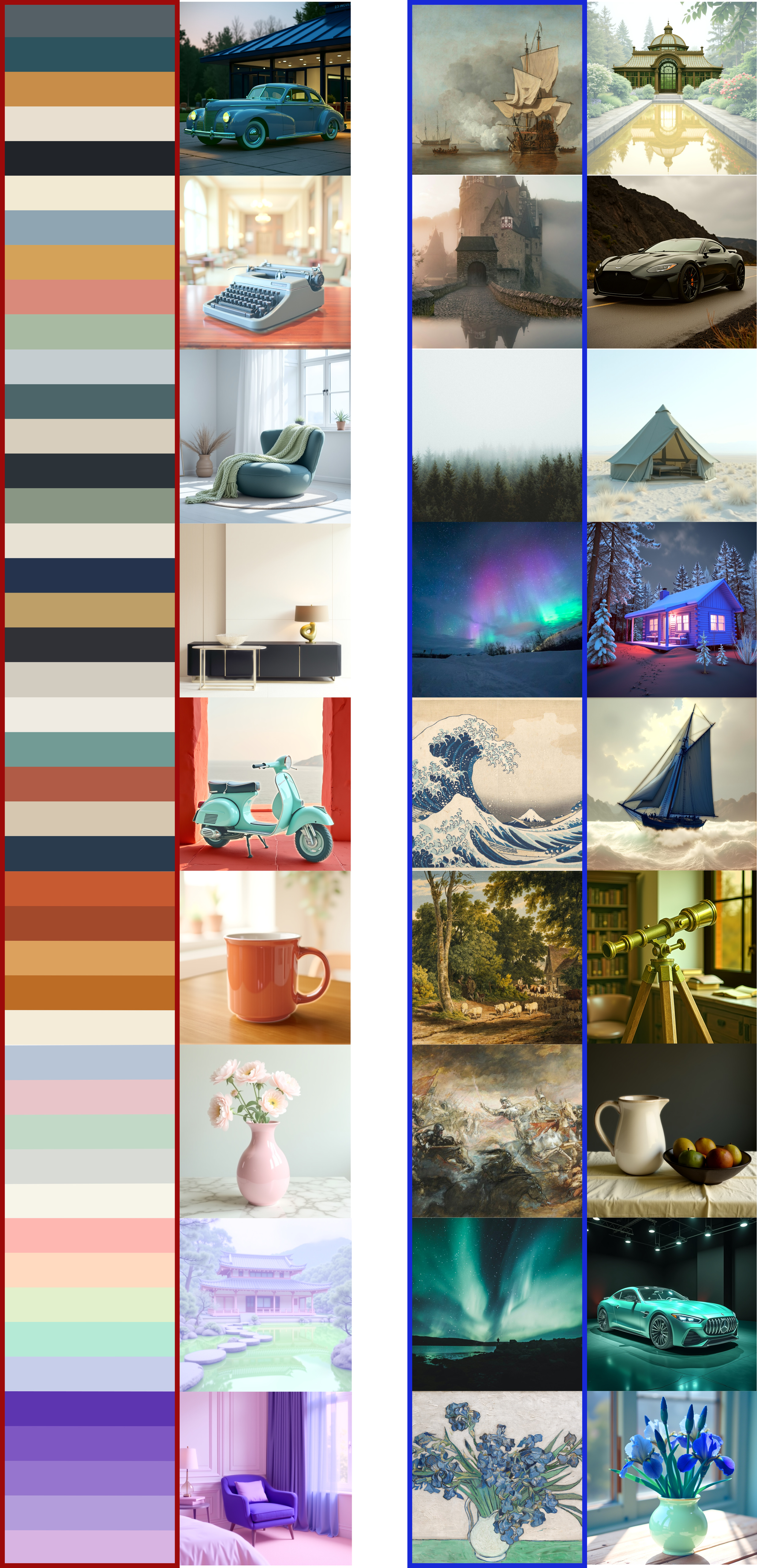}
    \caption{Color transfer results. The first two columns display the target color palettes and their corresponding generated images. The final two columns present the color reference images alongside the resulting generated outputs.}
    \label{fig:color_transfer}
\end{figure}

\textbf{Color Transfer.}
For color transfer, the target color distribution is specified either by a discrete color palette (e.g., a set of hex values) or by an exemplar reference image. When using an image, we first extract its principal colors via $k$-means clustering in CIELAB, enforcing a minimum perceptual distance in CIEDE2000 to obtain differentiated dominant, focal, and secondary accent tones. To initialize the trajectory in the appropriate chromatic basin, we take the most chromatic color, locate its nearest entry in ISCC-NBS Level 2, and inject this proxy name into the text prompt for the main subject. At gate step $s$, we decode the predicted clean latent $\hat{z}_0$ and partition the scene into spatial regions: the focal object segmented with SAM~3, secondary entities, and the background (which can be tonally partitioned into highlight and shadow zones). We assign each region to its corresponding reference color. The pretrained MLP then evaluates each pair of measured and target CIELAB colors independently, predicting a displacement $(m_1, m_2, m_3)$ for each region along the color basis. See qualitative results in Figure~\ref{fig:color_transfer}.
  
\subsection{User Study}\label{app:user_study}

\begin{figure}[http]
    \centering
    \includegraphics[width=0.8\linewidth]{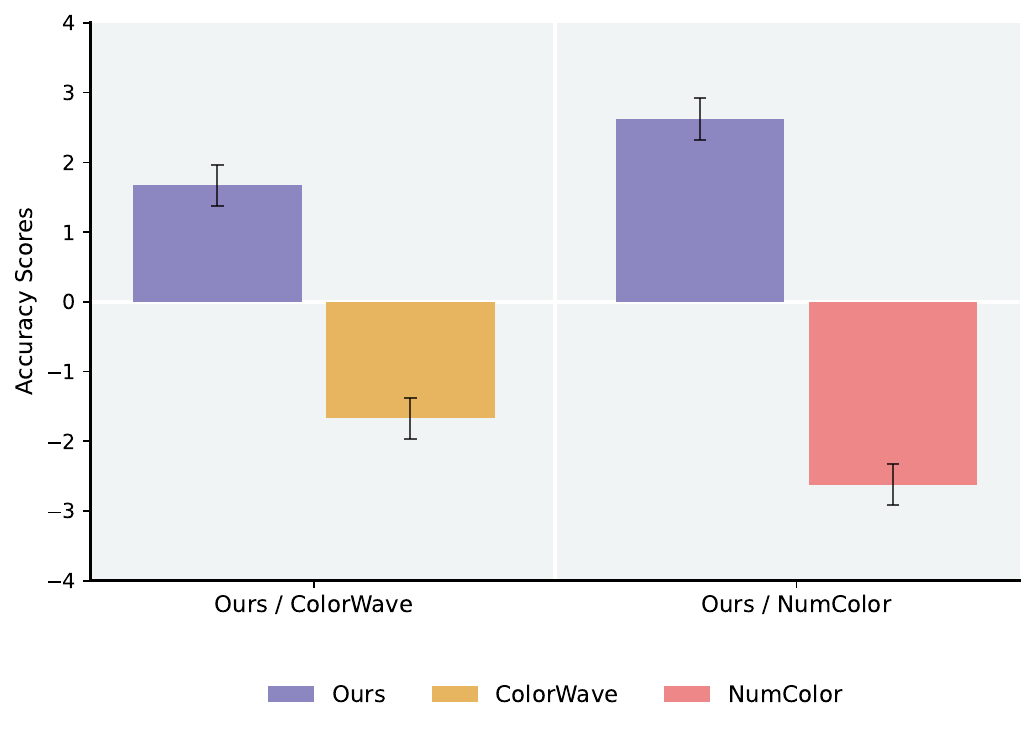}
    \caption{Thurstone case V results of our user’s study. Values are z-scores. Error bars represent 95\% confidence intervals —see \citet{montag2006empirical}. Our method is statistically better than existing methods: ColorWave and NumColor.}
    \label{fig:user_study_plot}
\end{figure}

We conducted a psychophysical experiment in a controlled laboratory with standardized lighting and color-calibrated sRGB monitors to perceptually evaluate our color accuracy. A total of 22 participants with normal color vision completed a Two-Alternative Forced Choice (2AFC) study consisting of two distinct comparisons based on the underlying diffusion backbone: \method against NumColor (both FLUX-based, 50 prompts) and \method against ColorWave (both SDXL-based, 50 prompts). In each trial, observers viewed a central reference color patch and text prompt positioned between two randomly ordered generated images (ours and the corresponding baseline), and were asked to choose which image better matched the target color. We analyzed the raw preference data using the Thurstone Case V Law of Comparative Judgment to compute accuracy z-scores, with 95\% confidence intervals calculated following \citet{montag2006empirical}. As illustrated in Figure~\ref{fig:user_study_plot}, a z-score greater than zero indicates a preference for our method. \method demonstrated a statistically significant preference over both competitors, achieving a z-score of $2.62 \pm 0.30$ against NumColor and $1.67 \pm 0.30$ against ColorWave.

\end{document}